\documentclass[letterpaper]{article} %
\usepackage{aaai25}  %
\usepackage{times}  %
\usepackage{helvet}  %
\usepackage{courier}  %
\usepackage[hyphens]{url}  %
\usepackage{graphicx} %
\usepackage{natbib}  %
\usepackage{caption} %
\usepackage{algorithm}
\usepackage{algorithmic}
\usepackage{mathtools}
\usepackage{subcaption}
\usepackage[toc,abbreviations]{glossaries-extra}
\usepackage{thmtools}
\usepackage[table]{xcolor} 
\declaretheorem[name=Example]{example}
\newtheorem{remark}{Remark}

\usepackage{amsmath,amsfonts,bm}
\newabbreviation{rl}{RL}{Reinforcement Learning}

\newabbreviation{rtg}{RTG}{reward-to-go}

\newabbreviation{gcrl}{GCRL}{Goal-Conditioned Reinforcement Learning}

\newabbreviation{mlp}{MLP}{Multi-Layer Perceptron}

\newabbreviation{dt}{DT}{Decision Transformer}

\newabbreviation{atamp}{ATAMP}{Adaptive Task Action Motion Planner}

\newabbreviation{gan}{GAN}{Generative Adversarial Network}

\newabbreviation{rvs}{RvS}{Reinforcement learning via Supervised Learning}

\newabbreviation{bc}{BC}{behaviour cloning}

\newabbreviation{gc}{GC}{Goal-Conditioning}

\newabbreviation{mdp}{MDP}{Markov Decision Processes}
\newabbreviation{ga}{GA}{Goal Augmentation}
\newabbreviation{ctmc}{CTMC}{Continuous Time Markov Chain}
\newabbreviation{tp}{TP}{Trajectory Planning}
\newabbreviation{ic}{IC}{Instruction Completion}
\newabbreviation{ap}{AP}{Adaptive Planning}
\newabbreviation{sac}{SAC}{Soft Actor Critic}

\newabbreviation{dfm}{DFM}{Discrete Flow Model}
\newabbreviation{ode}{ODE}{Ordinary Differential Equation}
\newabbreviation{mlm}{MLM}{Masked Language Model}

\newcommand{\sect}[1]{Section~\ref{#1}}

\newcommand{\eqn}[1]{eq.~\ref{#1}}
\newcommand{\fig}[1]{figure~\ref{#1}}
\newcommand{\tbl}[1]{Table~\ref{#1}}
\newcommand{\supp}[1]{Appendix~\ref{#1}}

\newcommand{\model}{GenPlan\xspace}
\newcommand{\baby}{BabyAI\xspace}
\newcommand{\film}{FiLM\xspace}
\newcommand{\meta}{MetaWorld\xspace}

\newcommand{\te}[1]{\texttt{#1}}

\def\eqref#1{equation~\ref{#1}}

\def\1{\bm{1}}

\def\va{{\bm{a}}}

\def\vg{{\bm{g}}}

\def\vo{{\bm{o}}}

\def\vs{{\mathbf{s}}}

\def\vx{{\bm{x}}}

\DeclareMathAlphabet{\mathsfit}{\encodingdefault}{\sfdefault}{m}{sl}
\SetMathAlphabet{\mathsfit}{bold}{\encodingdefault}{\sfdefault}{bx}{n}

\newcommand{\tnll}{\mathcal{L}_{\text{NLL}}\xspace}
\newcommand{\tce}{\mathcal{L}_{\text{CE}}\xspace}
\newcommand{\ta}{\mathcal{L}_{\va}\xspace}
\newcommand{\ts}{\mathcal{L}_{\vs}\xspace}
\newcommand{\tg}{\mathcal{L}_{\vg}\xspace}

\newcommand{\tpol}{{\pi_\theta}}
\newcommand{\tll}{\mathcal{L}_\pi(\theta)\xspace}

\newcommand{\relu}{\mathrm{ReLU}}

\newcommand{\x}{x}
\newcommand{\pdata}{p_{\mathrm{data}}}

\newcommand{\pnoise}{p_{\mathrm{noise}}}
\newcommand{\E}{\mathbb{E}}

\newcommand{\relurate}{R^*}

\newcommand{\dt}{\mathrm{d}t}

\newcommand{\Dt}{\Delta t}
\newcommand{\kdelta}[2]{\delta\!\left\{#1,#2\right\}}

\newcommand{\ptdt}{p_{t+\mathrm{d} t | t}}

\newcommand{\noisemarg}{p_{t|1}}

\newcommand{\denoise}{p_{1|t}}

\newcommand{\tok}{\texttt{tok}}
\newcommand{\ctx}{\text{ctx}}

\usepackage{xspace}
\usepackage{booktabs}
\usepackage{multirow} 
\usepackage{newfloat}
\usepackage{listings}
\DeclareCaptionStyle{ruled}{labelfont=normalfont,labelsep=colon,strut=off} %
\floatstyle{ruled}
\newfloat{listing}{tb}{lst}{}
\floatname{listing}{Listing}
\title{GenPlan: Generative Sequence Models as Adaptive Planners}
\author{
    Akash Karthikeyan\textsuperscript{\rm 1}, Yash Vardhan Pant\textsuperscript{\rm 1}
}
\affiliations{
    \textsuperscript{\rm 1}University of Waterloo, Canada\\

    \{a9karthi, yash.pant\}@uwaterloo.ca
}

\usepackage{bibentry}

\begin{document}

\maketitle

\begin{abstract}
Sequence models have demonstrated remarkable success in behavioral planning by leveraging previously collected demonstrations. 
However, solving multi-task missions remains a significant challenge, particularly when the planner must adapt to unseen constraints and tasks, such as discovering goals and unlocking doors. 
Such behavioral planning problems are challenging to solve due to: 
a) agents failing to adapt beyond the single task learned through their reward function, and 
b) inability to generalize to new environments, e.g., those with walls and locked doors, when trained only in planar environments.
Consequently, state-of-the-art decision-making methods are limited to missions where the required tasks are well-represented in the training demonstrations and can be solved within a short (temporal) planning horizon.  
To address this, we propose \model: a stochastic and adaptive planner that leverages discrete-flow models for generative sequence modeling, enabling sample-efficient exploration and exploitation. This framework relies on an iterative denoising procedure to generate a sequence of goals and actions.  
This approach captures multi-modal action distributions and facilitates goal and task discovery, thereby generalizing to out-of-distribution tasks and environments, i.e., missions not part of the training data. We demonstrate the effectiveness of our method through multiple simulation environments. Notably, \model outperforms state-of-the-art methods by over $10\%$ on adaptive planning tasks, where the agent adapts to multi-task missions while leveraging demonstrations from single-goal-reaching tasks. Our code is available at https://github.com/CL2-UWaterloo/GenPlan.
\end{abstract}

\section{Introduction}
\begin{figure}[tb]
    \centering
    \begin{subfigure}[t]{0.49\columnwidth}
        \centering
        \includegraphics[width=\columnwidth]{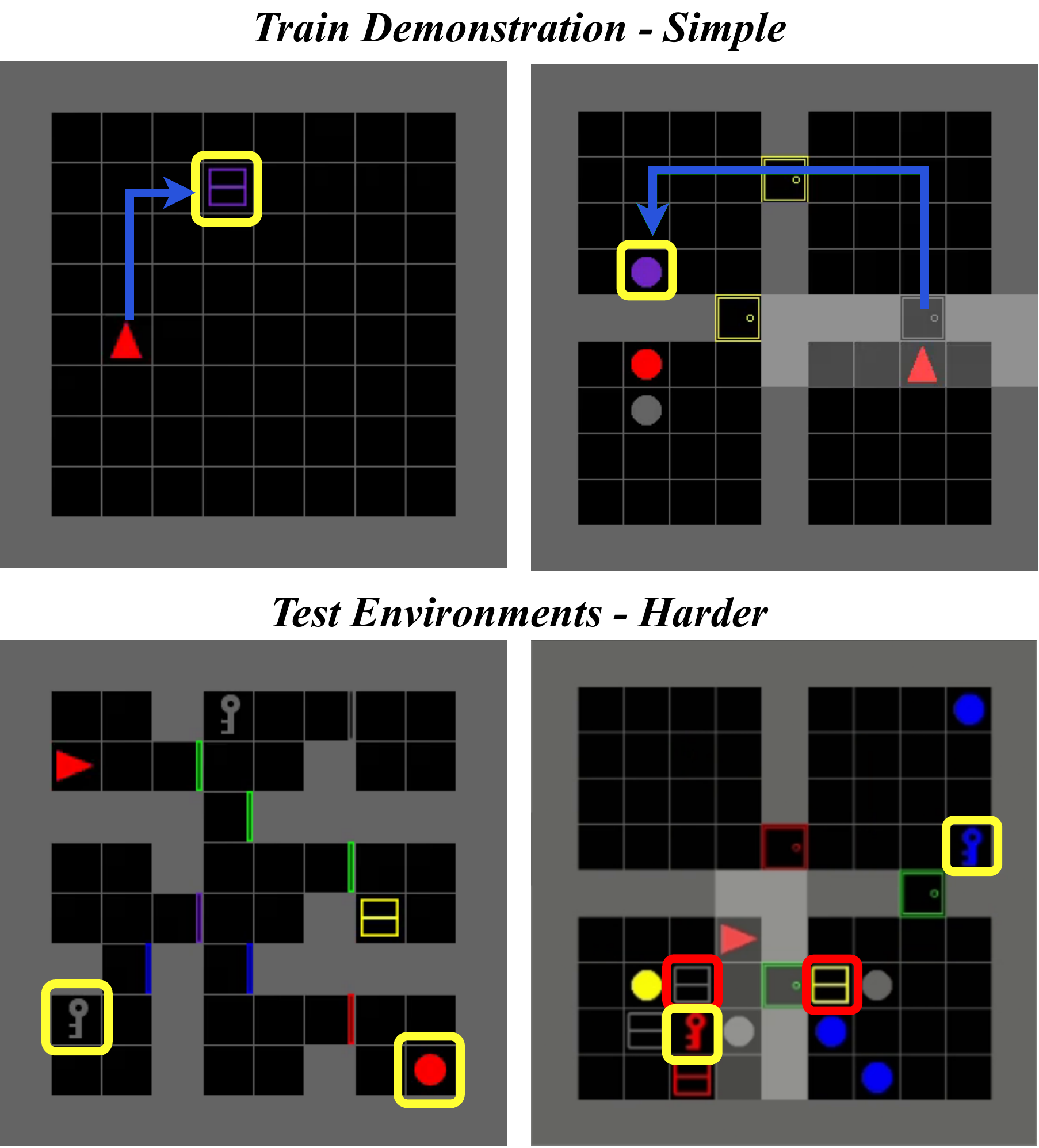}
        \caption{\textit{Adaptation to harder tasks.}}
        \label{fig:teaser_a}
    \end{subfigure}
    \hfill
    \begin{subfigure}[t]{0.49\columnwidth}
        \centering
        \includegraphics[width=\columnwidth]{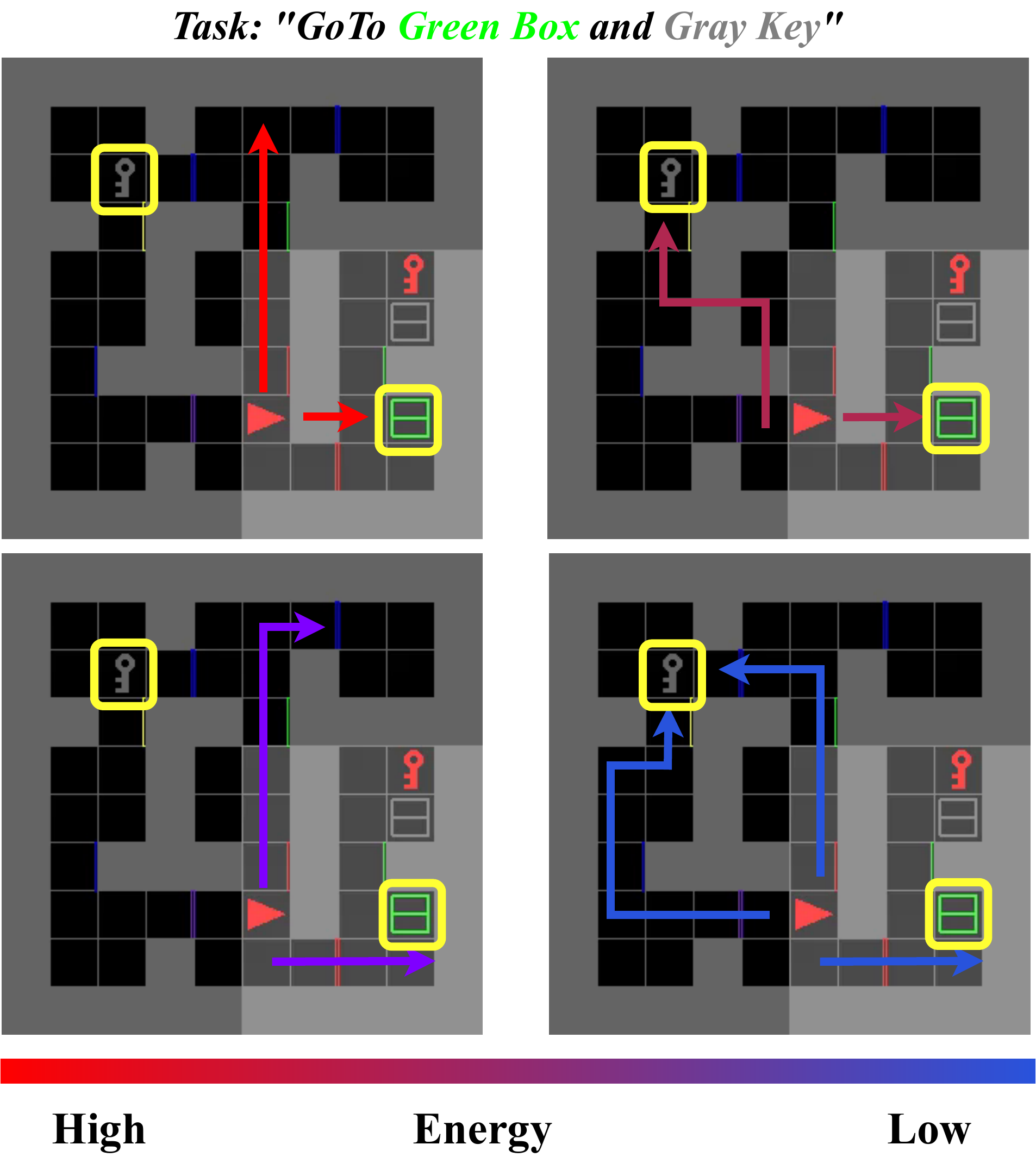}
        \caption{\textit{Iterative refinement.}}
        \label{fig:teaser_b}
    \end{subfigure}
    \caption{{\textbf{Overview.} \model is a generative, multi-step planner that optimizes energy landscape to adapt to complex tasks and iteratively refine long-horizon missions. Goals are highlighted in yellow, and distractors are marked in red.}}
    \label{fig:teaser}
\end{figure}

An intelligent autonomous agent must be adaptable to new
tasks at runtime beyond those encountered during training.
This is crucial for operating in complex environments that
may introduce distractors (i.e., objects the agent has not seen
before) and have multiple novel goals.

\begin{example}
\label{ex:adapt}
Consider a task where an agent is initially trained on a simple single-goal-reaching task in a planar environment. During evaluation, the agent must adapt to complex environments with walls, locked doors, and multiple goals, as illustrated in~\fig{fig:teaser}{A}. The planner must generate multi-step trajectories, exploring to locate goals while avoiding getting stuck, unblocking paths, and navigating around walls to complete the mission.
\end{example}
Learning such behaviors from demonstrations is challenging and often requires a diverse dataset that accurately represents the necessary tasks. This is much harder for long-horizon tasks, where the correct order of subtasks is critical (e.g., collecting a key before unlocking a door).

\gls{rvs}~\cite{emmons2022rvs} simplifies traditional \gls{rl} by using a 
\gls{bc} objective and framing \gls{rl} as sequence modeling, where actions 
are sequentially predicted based on next best action. However, this 
approach often struggles in order-critical multi-task missions, as generated 
plans can suffer from locally optimal decisions, resulting in global dead-ends, 
or accumulate errors without accounting for mistakes in previous 
actions~\cite{lambert2022compounding}.

Thus, a categorical distribution is better suited for capturing abstracted sequence planners~\cite{lee2024behavior}, as it enables clustering of different modes and provides interpretable predictions of subtask orders. More recently, generative modeling-based works allow conditional sequence-level predictions~\cite{janner2022planning, chi_diffusion_2023}. However, based on our results evaluating energy models and diffusion methods with various objectives (e.g., \gls{bc}) and sampling techniques, such as the cross-entropy method, Gibbs sampling, and energy gradient-based guidance, these approaches fail on multi-step planning (discrete) see \supp{sec:compare}. They often get stuck in local optima when handling unseen cases, as they lack distributional fidelity and struggle to adapt to new tasks.

\noindent{\textbf{Contributions of this work.}} To address these challenges, we propose \model, a discrete flow-based framework for sequence modeling that utilizes a \gls{ctmc}-based sampling method. Our key contributions are as follows:

\begin{itemize}
    \item \model frames planning as the iterative denoising of trajectories using discrete flow models. This approach enables adaptability by allowing goal and task discovery.
    \item Learn an energy function to guide denoising and minimize it to generate action sequences.
    \item \model uses an entropy-based lower bound over action probabilities to encourage adaptability and generalization to previously unseen tasks and environments.
\end{itemize}

Thus, instead of greedily selecting the next action, \model jointly learns goal and action distributions with bi-directional context. This approach prevents the agent from stalling and getting stuck in local regions.
Additionally, framing planning as generative modeling, \model allows for generalization to novel environments, as long as the objective function remains consistent (i.e., assigning minimal energy to successful trajectories).
Through extensive simulations, we empirically demonstrate that \model outperforms state-of-the-art models by over $10\%$ in adaptive planning, particularly in challenging environments with multiple sub-goals, while only leveraging demonstrations from single-goal-reaching tasks.

\section{Related Works}
\noindent\paragraph{Offline \gls{rl}.} Offline \gls{rl} focuses on learning policies from collected demonstrations, as illustrated in~\fig{fig:method}{A}, without further interaction with the environment~\cite{levine2020offline}. A key challenge in this approach is the \textit{distribution shift} between the training demonstrations and the runtime distribution. Several regularization strategies have been proposed to address this, such as reducing the discrepancy between the learned and behavioral policies~\cite{fujimoto2019off, kumar2019stabilizing}. However, these approaches often fail to adapt to unseen tasks and are typically limited to single-step planning. We are interested in optimizing sequence-level plans.

More recent works~\cite{chen_decision_2021, janner_offline_2021, furuta2021generalized} adopt an autoregressive modeling objective, leveraging the self-attention mechanism of sequence models. By conditioning on desired returns or goal states, these methods guide the generation of future actions, provided such states are encountered during training. While effective for behavior cloning tasks, they fail in scenarios like Example~\ref{ex:adapt} due to deadlocks. Moreover, they struggle in unconditional rollouts, lacking both guidance and the ability to generalize to novel goals.

\noindent\paragraph{Planning with Sequence Models.} 
Planning Transformer~\cite{plate} introduces procedural planning with 
transformers by framing planning as a model-based \gls{rl} problem. It employs 
a joint action and state representation model to mitigate compounding errors 
inherent in transformer-based architectures. LEAP~\cite{chen_planning_2023} 
addresses planning as an iterative energy minimization problem, utilizing a 
\gls{mlm} to learn trajectory-level energy functions. This approach employs 
Gibbs sampling prevents error accumulation and demonstrates generalization 
to novel test scenarios. However, LEAP's reliance on an oracle for goal 
positions limits its effectiveness, particularly in larger mazes, as shown 
in our simulation studies (see Table~\ref{tbl:performance_comparison}). 
Without conditional guidance, performance declines significantly. This limitation can cause the agent to enter loops, leading to \textit{stalling actions}, where it fails to make progress.

\noindent\paragraph{Generative Models in Planning.} 
Deep generative models have recently demonstrated success in offline \gls{rl}. Broadly, energy-based models~\cite{haarnoja2017reinforcement, eysenbach2022contrastive} and diffusion models~\cite{janner2022planning, chi_diffusion_2023} have been applied to offline \gls{rl} tasks. While diffusion-based models often require well-represented datasets, they fail to generalize to unseen harder tasks, as shown in~\fig{fig:teaser}A. In contrast, energy-based models can learn an energy landscape that helps generalize to unseen tasks; however, they are challenging to sample from, requiring cross-entropy or MCMC sampling~\cite{chen_planning_2023}, and are often unstable to train~\cite{chi_diffusion_2023}. Recently, energy-diffusion\cite{du2024learning} proposes guided diffusion sampling with energy gradients. However, in planning tasks, we observed that the low-energy trajectories can be gamed by repetitive frequent actions from the demonstrations (e.g., forward actions), causing the model to get trapped in local minima. 

To address these challenges, we propose an energy-diffusion framework (\model) capable of iteratively refining plans for unseen environments by learning annealed energy landscapes and employing diffusion-based sampling to enable goal and task discovery.

\section{Preliminaries and Problem Statement}
\label{sec:back}

\textbf{Notations.} We model discrete data as a sequence $(x_1, \dots, x_H)$, where $H$ is the horizon. This sequence is represented by $\vx$, with each $x_k$ denoting a step in the sequence. The superscript $x^t$ indicates the time step in the \gls{ctmc}, with $t \in [0, 1]$. Each $x^t_k \in X$ takes a discrete value from the set $X = \{1, \dotsc, x_{|X|}\}$, with $|X|$ denoting the cardinality of this set.
This notation applies to any discrete variable, such as a state \(s\) or an action \(a\). We denote the dataset by $\mathcal{D}$, representing a collection of demonstrations. We use $\mathcal{C}$ and $\mathcal{U}$ to denote categorical and uniform distributions, respectively. Additionally, we utilize $\kdelta{i}{j}$ to represent the Kronecker delta, which equals 1 when $i = j$ and 0 otherwise.

\subsubsection{Reinforcement Learning.} 
We extend the standard \gls{mdp} framework by incorporating a sequence of goals \(G = \{g_1, \dots, g_n\}\) within the state space \(S\). Formally, we consider learning in a \gls{mdp} $\mathcal{M} = \langle S, A, P, R \rangle$. The MDP tuple comprises $S$ and $A$ to denote the state and action spaces, respectively, with discrete states $s_k \in S$ and actions $a_k \in A$. The transition function $P(\cdot|s_k, a_k)$ returns a probability distribution over the next states given a state-action pair. 
The reward function $R$ provides a binary reward $r_k(s_k, a_k, s_{k+1}) = \mathbb{I}\{s_{k+1} \in G\}$. The aim is to learn a policy $\tpol(a_k|s_k)$ that maximizes cumulative rewards. While this approach allows for policy optimization, it is often restricted to single-step rollouts. In contrast, we focus on a multi-step planning framework that enables planning over a longer horizon, avoiding locally optimal decisions that may lead to global dead-ends.

\subsubsection{\gls{rl} as sequence modeling.}
\gls{dt}~\cite{chen_decision_2021} frames \gls{rl} as a sequence modeling problem, allowing training through upside-down \gls{rl}~\cite{schmidhuber2020reinforcement} in a supervised manner. Given a dataset $\mathcal{D} = \{\tau_i \mid 1 \leq i \leq N \}$ of near-optimal trajectories collected through demonstrations, where each trajectory $\tau_i = (s_1, a_1, g_1, \dotsc, s_H, a_H, g_H)$ has a length $H$, \gls{dt} aims to train an agent $\pi_\theta$ by minimizing the cross-entropy loss $\tce(\hat{\va}, \va)$ between the predicted actions $\hat{\va}$ and the true actions $\va$ in the demonstrations. Although this \gls{bc}-based approach has been successful in causal inference tasks (e.g., robotics), it lacks the ability for sequential refinement and is prone to error accumulation. Our approach addresses these limitations by integrating sequence models with an iterative denoising method for more efficient multi-step planning. 

\subsubsection{\gls{dfm}.} 
For simplicity, we assume $H=1$. However, this can be extended to multidimensional data by employing factorization assumptions, as demonstrated in~\cite{ campbell2024generative}.
Consider the probability flow $p_t$ as marginal distribution of $\x^t$ (samples in the \gls{ctmc}).
The objective of generative flow models is to transform source (noise) samples $p_{0}(\x^0) = \pnoise(\x^0)$ to target (data) samples $p_{1}(\x^1) = \pdata(\x^1)$. The probability that $\x^t$ will jump to other states $j$ is determined by a rate matrix $R_t \in \mathbb{R}^{|X| \times |X|}$. Thus, we can represent the transition probabilities for an infinitesimal time $\dt$ :
\begin{align}
    \label{eq:transition_prob}
    \ptdt (j | \x^t) &= \begin{cases}
        R_t(\x^t, j) \dt & \text{for} \,\, j \neq \x^t \\
        1 + R_t(\x^t, \x^t)\dt & \text{for} \,\, j=\x^t
        \end{cases} \\
        \label{eq:delta_fn}
        &= \kdelta{\x^t}{j}+ R_t(\x^t, j) \dt
\end{align}
where $\kdelta{i}{j}$ equals 1 when $i=j$ and 0 otherwise. We define a forward corruption process (data-to-noise interpolation) as $p_t(\x^t) = \E_{\pdata(\x^1)} \left[ \noisemarg(\x^t | \x^1) \right]$, with $p_t$ represented by the conditional flow $\noisemarg(\cdot | \x^1)$. We consider two types of interpolants: masking-based and uniform noise, both modeled as categorical distributions ($\mathcal{C}$):
\begin{subequations}
    \begin{align}
    \noisemarg^{\mathrm{mask}}(\x^t \mid \x^1) &= \mathcal{C}\left(t \kdelta{\x^1}{\x^t} + (1-t) \kdelta{[\texttt{M}]}{\x^t}\right) \label{eq:noise_mask} \\
    \noisemarg^{\mathrm{unif}}(\x^t \mid \x^1) &= \mathcal{C}\left(t \kdelta{\x^1}{\x^t} + (1-t) \frac{1}{|X|}\right) \label{eq:noise_unif}
    \end{align}
\end{subequations}
Here, $[\texttt{M}]$ represents a mask state~\cite{devlin2018bert} and $\frac{1}{|X|}$ denotes uniform prior.
To simulate the reverse flow (interpolation from noise to data), we need to determine the rate matrix $R$ that generates these marginals. This is expressed as $\partial_t p_t(\cdot)={R_t(\cdot|\cdot)^\top} p_t$. \citet{campbell2024generative} proposes the use of a conditional rate matrix, ${R_t(\x^t|j)} = \mathbb{E}_{\denoise} [{R_t(\x^t,j|\x^1)}]$, which enables the simulation of the conditional flow, by sampling along \gls{ctmc}.

\noindent\textbf{Learning a denoising model.} The $\denoise$ term is analytically intractable, but it can be approximated using a neural network with parameters $\theta$, and this model is commonly referred to as a denoising model $(\denoise^\theta)$ in literature. 

\subsubsection{Challenges.} 
Given an offline dataset, we wish to learn a planner that allows for sampling multi-step plans for a given task and environment. Primarily, we wish to address the following challenges: (1) Unconditional trajectory generation (goal position not available to the planner). (2) Even when goals are specified, goal-conditioned trajectories may provide insufficient guidance for long-horizon tasks, especially when the goal position is far from the agent's current position. (3) The agent may fail to adapt to multi-goal settings (harder) at test time, as \gls{bc}-based objectives result in memorization and are prone to accumulating errors over a long planning horizon. (4) Need for stochasticity in the planner, as we may encounter tasks previously unencountered (e.g., door-unlocking tasks) or unfamiliar environments (e.g., mazes).

\section{\model: Method}
\label{sec:method}

\begin{figure*}
    \centering
    \includegraphics[width=\linewidth]{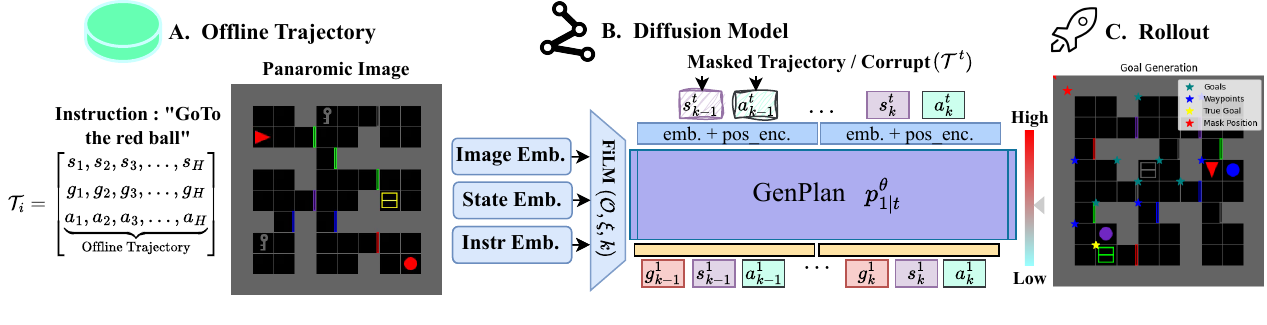} %
    \caption{\textbf{Method Overview.} \model, trained on offline data (\textbf{A}), learns to jointly model action, goal, and state distributions. In (\textbf{B}), the joint denoising model (see \sect{sec:joint_model}) takes in a corrupted trajectory $\tau^t$ and predicts the clean trajectory $\tau^1$. \textbf{(C)}~Demonstrates the joint inference of goals and actions by simulating the reverse \gls{ctmc}, as detailed in Algorithm~\ref{alg:sampling}.}
    \label{fig:method}
\end{figure*}

\noindent\textbf{Overview.} The denoising planner integrates the \gls{dfm} with sequence models and is built on two key components: (a) the denoising model $\denoise^\theta$, trained as described in algorithm~\ref{alg:training}, and (b) the rate matrix ${R_t(\vx^t,\vx'^{t}|\vx^1)}$. Together, these components enable flexible planning using the algorithm~\ref{alg:sampling}. We discuss several design choices that improve the planner's performance. For further details on the implementation and the formulation of the rate matrix, refer to~\supp{sec:rate_mat}.

\subsection{Energy-Guided Denoising Model}
\label{sec:energy}

Given demonstrations (see \sect{sec:back}), our aim is to learn an implicit energy function for a sequence, denoted as $\mathcal{E}(\va^t)$. This energy function is designed to assign lower energy to optimal action sequences. 
We define the energy as the sum of negative pseudo-likelihood over the horizon, formulated as $\mathcal{E}(\va^1) = \mathbb{E}_{(\va^1) \sim \mathcal{D}} \sum_{H} [-\log \denoise^\theta(\va^1 | \va^0, \vo)]$ adapted from~\cite{goyal2021exposing}.

Our approach leverages the \gls{dfm} objective~\cite{campbell2024generative, campbell2022continuous} to learn a locally normalized energy score. This score allows us to evaluate generated rollouts and frame planning as an iterative denoising process. Energy guidance ensures that the optimization process stays at an energy minimum at each step~\cite{goyal2021exposing, chen_planning_2023}. The denoiser $\denoise^\theta$ is optimized under the constraint that its entropy $\mathcal{H}$ remains above a lower bound $\beta$. This constraint encourages stochasticity, enhancing the generative model's adaptability to new tasks and environments.
\begin{subequations}
    \begin{align}
        \min_\theta &\ \mathbb{E}_{\va^0 \sim p_0, \vo \sim \mathcal{D}} \left[\sum_{{k}=1}^{{H}} -\log \denoise^\theta(\va^1 | \va^0, \vo) \right], \label{eqn:energy} \\
        \text{s.t.} &\ \mathbb{E}_{\va^0 \sim p_0, \vo \sim \mathcal{D}} \left[ \sum_{{k}=1}^{{H}} \mathcal{H}(\denoise^\theta(\va | \va^0,\vo)) \right] \geq \beta \label{eqn:constraint}
    \end{align}
\end{subequations}
By optimizing at the sequence level, the model avoids getting trapped in local minima and can naturally learn a multi-modal distribution, where $\mathcal{E}(\va) \simeq \mathcal{E}(\va')$, finding multiple viable solutions. 
This framework enables the flexible formulation of \gls{rl} and planning tasks. In \sect{sec:babyai_exps}, we demonstrate the use of a joint model (\sect{sec:joint_model}) for generating unconditional rollouts.

\subsection{Joint Denoising Model} 
\label{sec:joint_model}
\paragraph{Goal Generation.}
\model addresses complex planning tasks by using instruction prompts or the environment's observations to define objectives. We achieve this by learning a goal distribution conditioned on observations $\vo$, which guides the action denoising process. Unlike existing methods~\cite{chen_decision_2021, chen_planning_2023} that rely on simulators for runtime goal positions and task proposals, \model jointly learns the goal distribution, extracting a goal sequence $\vg$ jointly with actions $\va$ during planning. This dynamic goal proposal promotes exploration and improves performance in long-horizon tasks by reducing error accumulation.

\noindent\paragraph{State Sequence.} 
In addition to goal generation, we also learn a state denoising model. We found that goal conditioning becomes ambiguous when the proposed goal is far from the agent's position. To prevent the agent from stalling or getting stuck in local regions, we learn the state sequence, which further aids in learning the action distribution.

These auxiliary modules are trained similarly to the action sequence (denoising based \gls{dfm}) and can generalize to out-of-distribution tasks, provided the energy function is generalizable—successful trajectories receive low energy. Next, we briefly describe the processes involved in training and sampling from \gls{dfm}~\cite{campbell2024generative, campbell2022continuous}.

\noindent\paragraph{Forward Diffusion.} We begin by corrupting samples drawn from the dataset $\mathcal{D}$ using the noise schedules (\eqn{eq:noise_mask},\ref{eq:noise_unif}). Specifically, we apply $\E_{\pdata(\tau^1), t \sim \mathcal{U}[0, 1]} \noisemarg^{\mathrm{mask}}(\tau^t|\tau^1)$, where $t$ is the \gls{ctmc} timestep that controls the amount of corruption. The corruption process is applied to $\vs$, $\va$, and $\vg$, with the jump rate controlled by $t$. This training is independent of the rate matrix, offering greater flexibility during inference.

\noindent\paragraph{Backward Diffusion.} The denoising objective is applied at the trajectory level, making the transition from noise to data distribution challenging, particularly for longer horizons. This requires an iterative process. Using the corrupted tokens as input, we train the joint denoising model ($\denoise^\theta$) to approximate the true data distribution. We employ a negative-log-likelihood-based loss (\eqn{eq:general_loss}) to predict $\vs^1$, $\vg^1$, and $\va^1$. As noted in~\cite{campbell2024generative}, the denoiser $\denoise^\theta(\tau^1 \mid \tau^t)$ is independent of the choice of rate matrix used for sampling, ${R_t(\vx^t,\vx'^{t} \mid \vx^1)}$. This offers flexibility during sampling and simplifies the training. Following algorithm~\ref{alg:sampling}, we can sample the reverse \gls{ctmc} to generate rollouts.

\noindent\paragraph{Observation Conditioning.}
\label{sec:network}
We employ a transformer-style architecture~\cite{chen_decision_2021}, using bi-directional masks as in~\cite{devlin2018bert} to allow future actions to influence preceding ones. Observations $\vo$ are encoded using \film (Feature-wise Linear Modulation)~\cite{perez2018film} to process images $I$, instruction prompts $\xi$, and agent positions. The context length is typically set to 1 but can be extended to increase the agent's memory (see~\fig{fig:method}B).

\subsection{Training Objective}
\label{sec:training}
The training objectives described in~\eqn{eqn:energy},4b are used to learn the model parameters ($\theta, \lambda$). Here, $\theta$ represents the parameters of the denoiser $\denoise^\theta$, which is parameterized as a transformer, as discussed in~\sect{sec:network}. The parameter $\lambda \in [0, \infty]$ is the Lagrangian multiplier associated with the constraints in~\eqn{eqn:constraint}. We alternate gradient descent steps between updating $\denoise^\theta$ and $\lambda$. In practice, $\lambda \rightarrow 0$, ensuring that the lower bound $\beta$ on entropy is satisfied~\cite{zheng2022online}. To generalize the denoiser across different levels of corruption, we train the model to recover trajectories with varying noise levels. For each $\langle\va, \vs, \vg\rangle$, we apply the loss $\mathcal{L}_\vx$ (see~\eqn{eq:general_loss}). The Kronecker delta ensures that the loss is only calculated for the corrupted regions.

\begin{align}
    \mathcal{L}_\vx = \left[ -\sum_{k=1}^H \kdelta{\x_k^t}{[\texttt{M}]} \log \denoise^\theta(\vx^1 | \vx^t, \vo) \right]
    \label{eq:general_loss}
\end{align}

\begin{algorithm}[tp]
    \caption{\model Training}
    \label{alg:training}
    \begin{algorithmic}[1]
        \STATE \small{\textbf{init} denoiser $\denoise^\theta$, $\tau^0 \sim \pnoise$, $\beta=0.5$, $maxiters=5k$}
        \FOR{$i$ in ${maxiters}$}
        \STATE $t \sim \mathcal{U}[0,1],\quad \tau^1 \sim \mathcal{D}, \quad \vo \sim \mathcal{D}$
        \STATE $\noisemarg^{\mathrm{mask}}(\tau^t | \tau^1)$\small{ \te{ // see \eqn{eq:noise_mask}}}
        
        \STATE $\tnll \leftarrow \ta + \ts + \tg$ \small{ \te{ // see \eqn{eq:general_loss}}}
        \STATE {$\mathcal{L}_{\text{ent}}$ $\leftarrow$ $\mathbb{E}_{\va \sim \mathcal{D}} \left[\mathcal{H}\left(\denoise^\theta(\va \mid \tau^0, \vo)\right)\right]$}\small{ \te{ // see \eqn{eqn:constraint}}}
      
        \STATE ${\tll = \tnll - \lambda \mathcal{L}_{\text{ent}}}$ 
        \STATE $\theta \leftarrow  \theta - \nabla_\theta \tll$ 
        \STATE $\lambda \leftarrow \lambda - \left(\mathcal{H}\left[\denoise^\theta(\cdot \mid \tau^0, \vo)\right] - \beta\right)$
        \ENDFOR
    \end{algorithmic}
\end{algorithm}

\subsection{Planning}
\label{sec:planning}

Once the denoising model has been trained, we can generate trajectories that approximate the data distribution by reversing the \gls{ctmc} and interpolating noise back into the data through iterative denoising ($I_{max}$ iterations). The following algorithm outlines this process, based on the approach in \cite{campbell2024generative}. This procedure is applied to all components of the trajectories. 

\begin{algorithm}[H]
    \caption{\model Sampling}
    \label{alg:sampling}
    \begin{algorithmic}[1]
        \STATE \small{\textbf{init} $\tau^0 \sim p_0,\text{choice of } R_t(\tau^t,\cdot|\tau^1)$, $\Dt=\frac{1}{I_{max}}$, get $\vo$ }
        
        \FOR{$t \in \{0, \Dt, 2\Dt, \dotsc, 1\}$} 
            \STATE $R_t^\theta(\tau^t, \cdot) \gets \mathbb{E}_{p^\theta_{1|t}(\tau^1 | \tau^t, \vo) } \left[ R_t(\tau^t, \cdot|\tau^1) \right] $
            \STATE $\tau^{t+ \Dt} \sim \mathcal{C}\left(\kdelta{\tau^t}{\tau^{t+ \Dt}} + R_t^\theta(\tau^t, \tau^{t+ \Dt})  \Dt\right) $
            \STATE $t \gets t + \Dt$
        \ENDFOR
        \STATE \textbf{return} $\va, \vs, \vg$ \small{ \te{ // extract from $\tau^1$}}
    \end{algorithmic}
\end{algorithm}
\begin{remark}
     Extending rate matrices to sequences: For multidimensional input, line 4 of algorithm~\ref{alg:sampling} factorizes $\mathcal{C}(\cdot)$ as shown below. This is due to the following reasons: (1) The corruption process is independent across the horizon, meaning $R_t^k$ depends only on $x^t_k$, $j_k$, and $x^1_k$. (2) Once the process reaches $x^1_k$, it remains there, i.e., $R^k_t(x^t_k = x^1_k, j_k \mid x^1_k) = 0$.
\end{remark}

\begin{small}
\begin{equation}
    \delta \{ \vx^t, \vx'^t \} + \sum_{k=1}^H \delta \{ \vx^t_{\backslash k}, \vx'^t_{\backslash k} \} \E_{\denoise^\theta(\x^1_k | \vx^t) } \left[ R_t^k(\x^t_k, j_k | \x^1_k) \right] \Dt
    \nonumber
\end{equation}
\end{small}

\section{Simulation Studies}
We evaluate the performance of \model in \baby~\cite{babyai_iclr19} and continuous manipulation tasks, focusing on the agent's adaptive and generalization capabilities.
We implemented \model using Python 3.8 and trained it on a 12-core CPU alongside an RTX A6000 GPU.

\subsection{\baby}
\label{sec:babyai_exps}
\baby offers a diverse set of tasks and environments (discrete) focused on planning and task completion. 
\subsubsection{Simulation Setup.}
\label{sec:baby_setup}
We conduct simulations in a modified \baby suite following three paradigms.
\begin{enumerate}
\item \textbf{\gls{tp}.} The agent navigates to one or more goals in a maze world, with map layouts, agent initialization, and goal positions varied in each run, evaluating \model's generalization.
\item \textbf{\gls{ic}.} The agent operates in a multi-objective environment requiring complex decision-making, including \textit{exploration, object manipulation, key collection, and sequential goal completion}. Task order and navigation are critical.
\item \textbf{\gls{ap}.} The model, trained on simple goal-reaching tasks, is evaluated for zero-shot adaptation across harder environments without additional fine-tuning. This includes testing the agent in increasingly complex environments, such as mazes with multiple goals and closed doors, where the agent must demonstrate door-opening and navigation skills. Additionally, we assess the model's ability to adapt to environments in which obstacles must be unblocked to succeed.
\end{enumerate}
To navigate and interact with the environment, the agent can choose from six actions: \textit{left, right, forward, open, drop, or pick up}. Success rates in reaching goals and completing tasks are reported across 250 novel environments. The map layout, goals, obstacles, and agent positions are randomized in each run. As described in~\sect{sec:network}, the planner uses an image observation and instruction to denoise the action sequences. 
Further details on the environment, baseline configurations, and implementation are provided in~\supp{sec:app_babyai}.

\subsubsection{Baselines.}
We evaluate both versions of \model, named \model-M (GP-M) and \model-U (GP-U), corresponding to the masked and uniform noise variants, respectively. These are compared against several baselines:

\begin{enumerate}
    \item \textbf{LEAP}~\cite{chen_planning_2023}: A \gls{mlm} based multi-step planner, which has access to simulator-based goal-conditioning for sub-goals.
    \item \textbf{LEAP$\ominus$GC}: A variant of LEAP where simulator access is removed to evaluate its unconditional rollouts.
    \item \textbf{DT}~\cite{chen_decision_2021}: Utilizes a causal transformer for planning, also with simulator-based goal-conditioning.
\end{enumerate}

\noindent
While DT and LEAP could learn goal distributions, the MLE objective suffers from poor generalization, often predicting goal positions as corner cells. In contrast, the generative objective combined with DFM sampling in \model demonstrates superior generalization (\fig{fig:method}C). To improve competitiveness, baselines with simulator access to goal positions are categorized as conditional rollouts, while those relying on instructions and images to recover successful trajectories are categorized as unconditional rollouts. 
Comparative results are presented in Tables~\ref{tbl:performance_comparison} and~\ref{tbl:adaptive_comparison}.
\begin{remark}
LEAP outperforms popular baselines, including model-free \gls{rl} algorithms like Batch-Constrained Deep Q-Learning~\cite{fujimoto2019off} and Implicit Q-Learning~\cite{kostrikov2021offline}, as well as model-based \gls{rl} methods such as the Planning Transformer~\cite{plate} and Model-based Offline Policy~\cite{yu2020mopo}, in single-goal tasks within BabyAI environments, as shown in Table 1 of LEAP~\cite{chen_planning_2023}.
\end{remark}

\begin{table}[tp]
\centering
\small{
\setlength\tabcolsep{1mm} %
\renewcommand{\arraystretch}{1.2} %
\begin{tabular}{|l|c|c|c|c|c|}
    \hline
    \multicolumn{1}{|c|}{} & \multicolumn{3}{c|}{\textbf{Uncond. Rollouts}} & \multicolumn{2}{c|}{\textbf{Cond. Rollouts}} \\ 
    \cline{2-6}
    \multicolumn{1}{|c|}{\textbf{Env.}} & \textbf{GP-U} & \textbf{GP-M} & \textbf{LEAP$\ominus$GC} & \textbf{LEAP} & \textbf{DT} \\       
    \hline
    \multicolumn{6}{|c|}{\textbf{Traj. Planning (TP)}} \\ 
    \hline
    MazeS4G1 & {52.4\%} & \textbf{62\%} & 44\% & 49.2\% & 46.8\% \\ 
    MazeS4G2 & {38.8\%} & \textbf{39.6\%} & 20\% & 37.6\% & 35.2\% \\ 
    MazeS7G1 & \textbf{45.6\%} & {44.8\%} & 12\% & 33.2\% & 40\% \\ 
    MazeS7G2 & \textbf{21.2\%} & {19.6\%} & 3.6\% & 4\% & 13.6\% \\
    \hline
    \textbf{TP (7.6 $\uparrow$)} & 39.5\% & \textbf{41.5\%} & 19.9\% & 31\% & 33.9\% \\
    \hline
    \multicolumn{6}{|c|}{\textbf{Instr. Completion (IC)}} \\ 
    \hline
    MazeClose & {42.8\%} & \textbf{48.4\%} & 18\% & 38.8\% & 40\% \\
    DoorsOrder & \textbf{40.8\%} & 35.2\% & 11.2\% & 36.4\% & \textbf{40.8\%} \\
    BlockUn & 13.2\% & \textbf{16\%} & 0\% & 0.8\% & 0\% \\
    KeyCorS3R3 & {11.6\%} & \textbf{17.6\%} & 0\% & {0.4\%} & 3.6\% \\
    \hline
    \textbf{IC (8.2 $\uparrow$)} & 27.1\% & \textbf{29.3\%} & 7.25\% & 19.1\% & 21.1\% \\
    \hline
\end{tabular}}
\caption{{\textbf{\baby quantitative performance.} Success rates of the models across different environments are presented. The abbreviations \textbf{SW}, \textbf{NX}, \textbf{RY}, and \textbf{GZ} in the environment names represent the size (\textbf{W}) of a room in the map, the number of obstacles (\textbf{X}), the number of rows (\textbf{Y}), and the number of goals (\textbf{Z}) during testing, respectively. The term \textit{``Close''} indicates that the agent requires door-opening actions.}}
\label{tbl:performance_comparison}
\end{table}

\subsubsection{Results.}

\model consistently achieves higher success rates, particularly in adaptive planning and long-horizon tasks, as shown in \tbl{tbl:performance_comparison}. 
Success rates decline as environments grow or involve multiple goals (e.g., \textbf{\textit{MazeS7G2}}), reflecting the challenges of multi-step planning, where committing to incorrect plans results in deadlocks.
\model succeeds in complex, order-critical multi-task missions that require completing sub-tasks (e.g., key collection, obstacle unblocking) to achieve higher-level objectives, where baselines fail, particularly in \textbf{\textit{KeyCorS3R3}} and \textbf{\textit{BlockUn}}. In tasks requiring strict task order (e.g., \textbf{\textit{DoorsOrder}}), \gls{dt} performs comparably due to simulator access for correct goal sequencing, while \model relies solely on instruction embedding as seen in~\sect{sec:network}. 

\begin{figure}[tp]
    \centering
    \includegraphics[width=0.99\columnwidth]{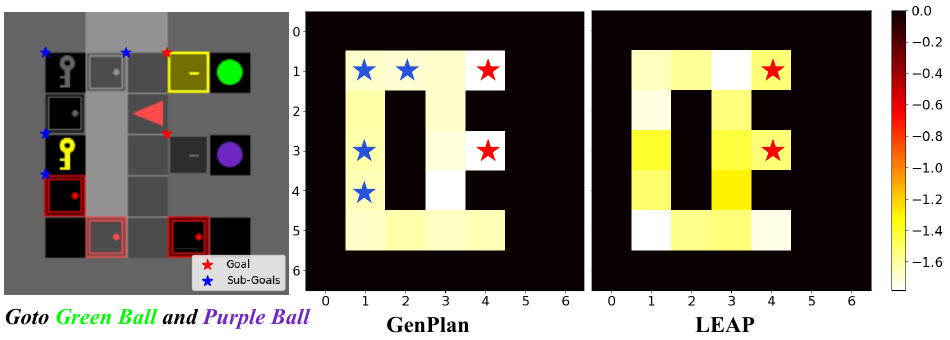} %
    \caption{{\textbf{Energy Landscape.} \model, when conditioned on sub-goals, implicitly assigns minimal energy to necessary sub-goals (e.g., picking up keys, opening doors) for task completion. States closer to the white region are more likely to be transitioned into. LEAP, in contrast, does not prioritize these sub-tasks.}}
    \label{fig:goal_cond}
\end{figure}

Next, we assess the model's adaptation capabilities in novel environments (see \tbl{tbl:adaptive_comparison}). Initially trained in a simple, single-goal plane world task, the model is then evaluated on increasingly challenging tasks. Through the joint denoising process, \model can eventually find a plan given enough timesteps as long as the energy function generalizes well. The goal-generation module allows dynamic updates of sub-goals, aiding exploration, while the entropy regularizer helps the model learn new skills, such as obstacle unblocking, even without explicit demonstrations in the dataset.
We hypothesize that this is due to \model's more effective joint optimization process. In practice, LEAP's sampling process may lead to misalignment between generated actions and goals, even with an oracle for the goal sequence (see~\fig{fig:goal_cond}).

\begin{table}[tp]
    \centering
    \setlength\tabcolsep{1mm} %
    \small{
    \begin{tabular}{|l|c|c|c|c|c|}
        \hline
         & \multicolumn{3}{c|}{\textbf{Uncond. Rollouts}} & \multicolumn{2}{c|}{\textbf{Cond. Rollouts}} \\
        \hline
        \textbf{Environment} & \textbf{GP-U} & \textbf{GP-M} & \textbf{LEAP$\ominus$GC} & \textbf{LEAP} & \textbf{DT}\\
        \hline
        \multicolumn{6}{|c|}{\textbf{Adaptive Planning (AP)}} \\
        \hline
        LocS10N10G2 & 82.4\%  & \textbf{88\%} & 76\% & 78\% & 25.6\%\\
        MazeS4N3G1 & 56\% & \textbf{62\%} & 44.8\% & 48\% & 24\% \\ 
        MazeClose & 31.2\% & \textbf{34.8\%} & 10\% & 10\% & 8.8\% \\
        MazeS4G2 & 28.8\% & \textbf{34.8\%} & 14\% & 18.4\% & 3.6\% \\ 
        SeqS5R2Un & 35.6\% & \textbf{42\%} & 29.2\% & 38\% & 29.2\% \\
        \hline
        \textbf{AP (13.84 $\uparrow$)} & 46.8\% & \textbf{52.3\%} & 34.8\% & 38.4\% & 18.2\% \\
        \hline
    \end{tabular}}
    \caption{{\textbf{\baby quantitative performance.} Success rates of the models across different environments are presented. The abbreviations \textbf{SW}, \textbf{NX}, \textbf{RY}, and \textbf{GZ} in the environment names represent the size (\textit{W}) of a room in the map, the number of obstacles (\textbf{X}), the number of rows (\textbf{Y}), and the number of goals (\textbf{Z}) during testing, respectively.}}
    \label{tbl:adaptive_comparison}
\end{table}
\subsubsection{Ablation Studies.} We assess the significance of various components of \model, results are presented in~\tbl{tbl:ablation}. 

\textbf{Joint Prediction ($\ts + \tg$).} This integration of goal generation ($\tg$) with state sequence denoising ($\ts$) enhances \gls{ic} performance and prevents stalling in \gls{tp} tasks. Without state denoising, the agent tends to exhibit stalling actions as the agent may get stuck in a local region. On the other hand, without $\tg$, the agent lacks awareness of sub-tasks associated with the overall task. This is evident in the performance drops observed in the \textbf{\textit{KeyCorS3R3}} and \textbf{\textit{DoorsOrder}} experiments in~\tbl{tbl:ablation}, where the agent struggles to achieve sub-goals like key collection or correctly identifying the door sequence.

\textbf{Entropy.} Entropy acts as a regularizer while minimizing the trajectory's energy. LEAP often gets stuck in local minima when recovering trajectories, leading to the hallucination of actions from demonstrations, such as constantly moving forward or performing in-place turns. The impact of entropy is particularly significant in \gls{ap} tasks, where the agent navigates more complex environments. This is evident in the performance drops observed in the \textbf{\textit{MazeClose}} environment. However, in \gls{ic} tasks, where additional actions like ``\textit{open}'' or ``\textit{pickup}'' are required, entropy often hinders performance, making it less suitable for such scenarios.

\textbf{Noise Schedule.} In Tables~\ref{tbl:performance_comparison} and~\ref{tbl:adaptive_comparison}, we compare the use of uniform interpolants versus masking. While masking interpolants generally outperform uniform interpolants in most tasks, the latter succeed in some larger environments. This success is attributed to the inherent randomness in the denoising process, as discussed in~\sect{sec:planning}. With masking, the model more easily identifies jumps, simplifying the training. This is also reflected through faster convergence (training) of masked interpolants.

\begin{table}[ht]
    \centering
    \setlength\tabcolsep{1mm} %
    \small{
    \begin{tabular}{l r r r}
        \toprule
        \textbf{Attribute} & \textbf{\model-M} & \textbf{Reduction} & \textbf{LEAP$\ominus$GC}  \\
        \midrule
        \multicolumn{4}{c}{\textbf{MazeS4G2} (\gls{tp})} \\
        w/o JP & \textbf{32\%} & {\textbf{$\downarrow$ 7.6\%}} & \multirow{2}{*}{20\%} \\
        w/o Entropy & \textbf{34.4\%} & {\textbf{$\downarrow$ 5.2\%}} & \\
        
        \multicolumn{4}{c}{\textbf{KeyCorS3R3}  (\gls{ic})} \\
        w/o JP & \textbf{13.6\%} & {\textbf{$\downarrow$ 4\%}} & \multirow{2}{*}{0\%} \\
        w/o Entropy & \textbf{7.6\%} & {\textbf{$\downarrow$ 10\%}} & \\
        w/o History & \textbf{0.4\%} & {\textbf{$\downarrow$ 17.2\%}} & 0\%\\
        
        \multicolumn{4}{c}{\textbf{DoorsOrder}  (\gls{ic})} \\
        w/o JP & \textbf{15.6\%} & {\textbf{$\downarrow$ 19.6\%}} & \multirow{2}{*}{11\%}\\
        w/o Entropy & \textbf{34.4\%} & {\textbf{$\downarrow$ 0.8\%}} & \\

         \multicolumn{4}{c}{\textbf{SeqS5R2Un}  (\gls{ap})}\\
         w/o JP & \textbf{32.8\%} & {\textbf{$\downarrow$ 9.2\%}} & \multirow{2}{*}{29.2\%}\\
        w/o Entropy & \textbf{34.8\%} & {\textbf{$\downarrow$ 7.2\%}} & \\
        
        \bottomrule
        
    \end{tabular}}
    \caption{{\textbf{Ablation.} The values represent the mean success rates as 
detailed in~\sect{sec:baby_setup}, with reductions shown relative to the results 
in Tables~\ref{tbl:performance_comparison} and~\ref{tbl:adaptive_comparison}.}}
    \label{tbl:ablation}
\end{table}
\subsubsection{Discussion.}  
Unlike baselines that rely on an oracle for goal positions—often unavailable in real-world scenarios—the goal generation module in \model dynamically updates sub-goals, enabling better state coverage (see~\supp{sec:app_babyai}) and unconditional rollouts, where baselines fail to explore certain regions even with oracle guidance. To further illustrate these advantages, we present results on training progression through the learned energy landscape, mission success rates as a function of iterations, and the impact of environment stochasticity (see~\supp{sec:app_babyai}).

\subsection{Continuous Tasks}

We evaluate the models on continuous manipulation tasks: (a) PushT~\cite{florence2021implicit} and (b) Franka Kitchen~\cite{gupta2019relay}. Both tasks use state-based variants in continuous manipulator environments. The PushT task assesses task comprehension in unconditional rollouts, with performance measured by final coverage, defined as the IoU between the T-block and the target position. The Franka Kitchen task involves seven possible subtasks, where each demonstration trajectory completes a subset of four in some order. It evaluates long-horizon planning, multimodality, and task commitment, where certain baselines may fail due to mode-shifting or incomplete tasks. Both tasks are evaluated in an unconditional rollout setting. For details on the environment setup and implementation, see~\supp{sec:cont_tasks}.

\subsubsection{Baselines.}
For our evaluations, we compare \model-M with several baselines to highlight its performance in continuous tasks. \model-M retrieves continuous actions by learning a categorical distribution over discrete latent codes, with corresponding offsets, similar to the Vector Quantized Behavior Transformer (VQ-BeT)~\cite{lee2024behavior}. 

Unlike VQ-BeT, which employs a behavior cloning objective, \model-M uses an energy-based loss (see \sect{sec:method}) and DFM sampling with casual masks, providing greater flexibility and improved generalization. Additionally, we benchmark against Diffusion Policy (DP), a state-of-the-art behavior cloning method, evaluating both its CNN-based (DP-C) and Transformer-based (DP-T) variants. For model and environment hyperparameters, we adopt the configurations from~\cite{lee2024behavior}.

\subsubsection{Results.} 
\model-M demonstrates competitive performance against state-of-the-art behavior cloning approaches. While diffusion policies excel at low-level tasks such as PushT, \model-M, and VQ-BeT outperforms the diffusion-based approach (DP-C) in the Kitchen environment, where diffusion falls behind by an entire subtask. These results highlight the advantages of modeling with categorical distributions for multi-task missions and unconditional planning.

\section{Conclusion}
\begin{table}[tb]
\centering
\setlength\tabcolsep{1mm} %
\begin{small}
\scalebox{1.0}{
\begin{tabular}{|l|l|c|c|c|c|}
\hline
\textbf{Env} & \textbf{Metric} & \textbf{\model-M} & \textbf{DP-C} & \textbf{DP-T} & \textbf{VQ-BeT} \\ \hline
PushT        & IoU             & 0.73              & 0.73          & \textbf{0.74} & 0.68           \\ \hline
Kitchen      & \# Tasks        & 3.40               & 2.62          & 3.44          & \textbf{3.66}  \\ \hline
\end{tabular}
}
\end{small}
\caption{\textbf{Quantitative results.} Comparison of baselines on unconditional continuous-space tasks.}
\label{table:env_metrics_comparison}
\end{table}

We study the problem of learning to plan from demonstrations, particularly for unseen tasks and environments. We propose \model, an energy-\gls{dfm}-based planner that learns annealed energy landscapes and uses \gls{dfm} sampling to iteratively denoise plans. Through simulation studies, we demonstrate how joint energy-based denoising improves performance in complex and long-horizon tasks.

\noindent\paragraph{Limitations.}
(1) The entropy lower bound $\beta$ in~\eqn{eqn:constraint} is currently a hyper-parameter that must be manually specified. (2) We assume access to near-optimal demonstration trajectories for training (\sect{sec:babyai_exps}); however, this assumption may not hold in all settings. Initial results show that \model performs well even when trained on datasets with a mixture of sub-optimal demonstrations (\supp{sec:meta_data}). However, further studies are needed to assess its robustness to sub-optimality in demonstrations.

\noindent\paragraph{Future Work.} 
To address the above limitations, we plan to extend \model for online fine-tuning via hindsight experience replay~\cite{zheng2022online, furuta2021generalized}. Additionally, \model offers a flexible and scalable framework that can be extended to multi-agent learning settings~\cite{meng2022offlinepretrainedmultiagentdecision}.

\section*{Acknowledgments}
This research is supported by the Natural Sciences and Engineering Research Council of Canada (NSERC) and by the Vector Scholarship in Artificial Intelligence, provided through the Vector Institute. 

\bibliography{aaai25}

\begin{thebibliography}{31}
\providecommand{\natexlab}[1]{#1}

\bibitem[{Campbell et~al.(2022)Campbell, Benton, De~Bortoli, Rainforth, Deligiannidis, and Doucet}]{campbell2022continuous}
Campbell, A.; Benton, J.; De~Bortoli, V.; Rainforth, T.; Deligiannidis, G.; and Doucet, A. 2022.
\newblock A continuous time framework for discrete denoising models.
\newblock \emph{Advances in Neural Information Processing Systems}.

\bibitem[{Campbell et~al.(2024)Campbell, Yim, Barzilay, Rainforth, and Jaakkola}]{campbell2024generative}
Campbell, A.; Yim, J.; Barzilay, R.; Rainforth, T.; and Jaakkola, T. 2024.
\newblock Generative Flows on Discrete State-Spaces: Enabling Multimodal Flows with Applications to Protein Co-Design.
\newblock arXiv:2402.04997.

\bibitem[{Chen et~al.(2023)Chen, Du, Chen, Tenenbaum, and Vela}]{chen_planning_2023}
Chen, H.; Du, Y.; Chen, Y.; Tenenbaum, J.~B.; and Vela, P.~A. 2023.
\newblock Planning with Sequence Models through Iterative Energy Minimization.
\newblock In \emph{{ICLR}}.

\bibitem[{Chen et~al.(2021)Chen, Lu, Rajeswaran, Lee, Grover, Laskin, Abbeel, Srinivas, and Mordatch}]{chen_decision_2021}
Chen, L.; Lu, K.; Rajeswaran, A.; Lee, K.; Grover, A.; Laskin, M.; Abbeel, P.; Srinivas, A.; and Mordatch, I. 2021.
\newblock Decision Transformer: Reinforcement Learning via Sequence Modeling.
\newblock In \emph{Advances in Neural Information Processing Systems}, volume~34, 15084--15097.

\bibitem[{Chevalier-Boisvert et~al.(2019)Chevalier-Boisvert, Bahdanau, Lahlou, Willems, Saharia, Nguyen, and Bengio}]{babyai_iclr19}
Chevalier-Boisvert, M.; Bahdanau, D.; Lahlou, S.; Willems, L.; Saharia, C.; Nguyen, T.~H.; and Bengio, Y. 2019.
\newblock Baby{AI}: First Steps Towards Grounded Language Learning With a Human In the Loop.
\newblock In \emph{International Conference on Learning Representations}.

\bibitem[{Chi et~al.(2023)Chi, Feng, Du, Xu, Cousineau, Burchfiel, and Song}]{chi_diffusion_2023}
Chi, C.; Feng, S.; Du, Y.; Xu, Z.; Cousineau, E.; Burchfiel, B.; and Song, S. 2023.
\newblock Diffusion {Policy}: {Visuomotor} {Policy} {Learning} via {Action} {Diffusion}.
\newblock ArXiv:2303.04137 [cs].

\bibitem[{Devlin et~al.(2018)Devlin, Chang, Lee, and Toutanova}]{devlin2018bert}
Devlin, J.; Chang, M.-W.; Lee, K.; and Toutanova, K. 2018.
\newblock Bert: Pre-training of deep bidirectional transformers for language understanding.
\newblock \emph{arXiv preprint arXiv:1810.04805}.

\bibitem[{Du, Mao, and Tenenbaum(2024)}]{du2024learning}
Du, Y.; Mao, J.; and Tenenbaum, J.~B. 2024.
\newblock Learning Iterative Reasoning through Energy Diffusion.
\newblock In \emph{Forty-first International Conference on Machine Learning}.

\bibitem[{Emmons et~al.(2022)Emmons, Eysenbach, Kostrikov, and Levine}]{emmons2022rvs}
Emmons, S.; Eysenbach, B.; Kostrikov, I.; and Levine, S. 2022.
\newblock RvS: What is Essential for Offline {RL} via Supervised Learning?
\newblock In \emph{International Conference on Learning Representations}.

\bibitem[{Eysenbach et~al.(2022)Eysenbach, Zhang, Levine, and Salakhutdinov}]{eysenbach2022contrastive}
Eysenbach, B.; Zhang, T.; Levine, S.; and Salakhutdinov, R. 2022.
\newblock Contrastive Learning as Goal-Conditioned Reinforcement Learning.
\newblock In Oh, A.~H.; Agarwal, A.; Belgrave, D.; and Cho, K., eds., \emph{Advances in Neural Information Processing Systems}.

\bibitem[{Florence et~al.(2021)Florence, Lynch, Zeng, Ramirez, Wahid, Downs, Wong, Lee, Mordatch, and Tompson}]{florence2021implicit}
Florence, P.; Lynch, C.; Zeng, A.; Ramirez, O.; Wahid, A.; Downs, L.; Wong, A.; Lee, J.; Mordatch, I.; and Tompson, J. 2021.
\newblock Implicit Behavioral Cloning.
\newblock \emph{Conference on Robot Learning (CoRL)}.

\bibitem[{Fujimoto, Meger, and Precup(2019)}]{fujimoto2019off}
Fujimoto, S.; Meger, D.; and Precup, D. 2019.
\newblock Off-Policy Deep Reinforcement Learning without Exploration.
\newblock In \emph{International Conference on Machine Learning}, 2052--2062.

\bibitem[{Furuta, Matsuo, and Gu(2022)}]{furuta2021generalized}
Furuta, H.; Matsuo, Y.; and Gu, S.~S. 2022.
\newblock Generalized Decision Transformer for Offline Hindsight Information Matching.
\newblock In \emph{International Conference on Learning Representations}.

\bibitem[{Goyal, Dyer, and Berg-Kirkpatrick(2021)}]{goyal2021exposing}
Goyal, K.; Dyer, C.; and Berg-Kirkpatrick, T. 2021.
\newblock Exposing the Implicit Energy Networks behind Masked Language Models via Metropolis--Hastings.
\newblock \emph{arXiv preprint arXiv:2106.02736}.

\bibitem[{Gupta et~al.(2019)Gupta, Kumar, Lynch, Levine, and Hausman}]{gupta2019relay}
Gupta, A.; Kumar, V.; Lynch, C.; Levine, S.; and Hausman, K. 2019.
\newblock Relay policy learning: Solving long-horizon tasks via imitation and reinforcement learning.
\newblock \emph{arXiv preprint arXiv:1910.11956}.

\bibitem[{Haarnoja et~al.(2017)Haarnoja, Tang, Abbeel, and Levine}]{haarnoja2017reinforcement}
Haarnoja, T.; Tang, H.; Abbeel, P.; and Levine, S. 2017.
\newblock Reinforcement Learning with Deep Energy-Based Policies.
\newblock arXiv:1702.08165.

\bibitem[{Ho, Jain, and Abbeel(2020)}]{ho2020denoising}
Ho, J.; Jain, A.; and Abbeel, P. 2020.
\newblock Denoising Diffusion Probabilistic Models.
\newblock arXiv:2006.11239.

\bibitem[{Janner et~al.(2022)Janner, Du, Tenenbaum, and Levine}]{janner2022planning}
Janner, M.; Du, Y.; Tenenbaum, J.~B.; and Levine, S. 2022.
\newblock Planning with Diffusion for Flexible Behavior Synthesis.
\newblock arXiv:2205.09991.

\bibitem[{Janner, Li, and Levine(2021)}]{janner_offline_2021}
Janner, M.; Li, Q.; and Levine, S. 2021.
\newblock Offline {Reinforcement} {Learning} as {One} {Big} {Sequence} {Modeling} {Problem}.

\bibitem[{Kostrikov, Nair, and Levine(2021)}]{kostrikov2021offline}
Kostrikov, I.; Nair, A.; and Levine, S. 2021.
\newblock Offline reinforcement learning with implicit q-learning.
\newblock \emph{arXiv preprint arXiv:2110.06169}.

\bibitem[{Kumar et~al.(2019)Kumar, Fu, Soh, Tucker, and Levine}]{kumar2019stabilizing}
Kumar, A.; Fu, J.; Soh, M.; Tucker, G.; and Levine, S. 2019.
\newblock Stabilizing off-policy q-learning via bootstrapping error reduction.
\newblock \emph{Advances in Neural Information Processing Systems}, 32.

\bibitem[{Lambert, Pister, and Calandra(2022)}]{lambert2022compounding}
Lambert, N.; Pister, K.; and Calandra, R. 2022.
\newblock Investigating Compounding Prediction Errors in Learned Dynamics Models.
\newblock arXiv:2203.09637.

\bibitem[{Lee et~al.(2024)Lee, Wang, Etukuru, Kim, Shafiullah, and Pinto}]{lee2024behavior}
Lee, S.; Wang, Y.; Etukuru, H.; Kim, H.~J.; Shafiullah, N. M.~M.; and Pinto, L. 2024.
\newblock Behavior Generation with Latent Actions.
\newblock \emph{arXiv preprint arXiv:2403.03181}.

\bibitem[{Levine et~al.(2020)Levine, Kumar, Tucker, and Fu}]{levine2020offline}
Levine, S.; Kumar, A.; Tucker, G.; and Fu, J. 2020.
\newblock Offline Reinforcement Learning: Tutorial, Review, and Perspectives on Open Problems.
\newblock arXiv:2005.01643.

\bibitem[{Meng et~al.(2022)Meng, Wen, Yang, Le, Li, Zhang, Wen, Zhang, Wang, and Xu}]{meng2022offlinepretrainedmultiagentdecision}
Meng, L.; Wen, M.; Yang, Y.; Le, C.; Li, X.; Zhang, W.; Wen, Y.; Zhang, H.; Wang, J.; and Xu, B. 2022.
\newblock Offline Pre-trained Multi-Agent Decision Transformer: One Big Sequence Model Tackles All SMAC Tasks.
\newblock arXiv:2112.02845.

\bibitem[{Perez et~al.(2018)Perez, Strub, De~Vries, Dumoulin, and Courville}]{perez2018film}
Perez, E.; Strub, F.; De~Vries, H.; Dumoulin, V.; and Courville, A. 2018.
\newblock Film: Visual reasoning with a general conditioning layer.
\newblock In \emph{Proceedings of the AAAI Conference on Artificial Intelligence}.

\bibitem[{Schmidhuber(2020)}]{schmidhuber2020reinforcement}
Schmidhuber, J. 2020.
\newblock Reinforcement Learning Upside Down: Don't Predict Rewards -- Just Map Them to Actions.
\newblock arXiv:1912.02875.

\bibitem[{Schmied et~al.(2023)Schmied, Hofmarcher, Paischer, Pascanu, and Hochreiter}]{schmied2023learning}
Schmied, T.; Hofmarcher, M.; Paischer, F.; Pascanu, R.; and Hochreiter, S. 2023.
\newblock Learning to Modulate pre-trained Models in {RL}.
\newblock In \emph{Thirty-seventh Conference on Neural Information Processing Systems}.

\bibitem[{Sun et~al.(2022)Sun, Huang, Lu, Liu, Zhou, and Garg}]{plate}
Sun, J.; Huang, D.-A.; Lu, B.; Liu, Y.-H.; Zhou, B.; and Garg, A. 2022.
\newblock PlaTe: Visually-Grounded Planning With Transformers in Procedural Tasks.
\newblock \emph{IEEE Robotics and Automation Letters}, 7(2): 4924--4930.

\bibitem[{Yu et~al.(2020)Yu, Thomas, Yu, Ermon, Zou, Levine, Finn, and Ma}]{yu2020mopo}
Yu, T.; Thomas, G.; Yu, L.; Ermon, S.; Zou, J.; Levine, S.; Finn, C.; and Ma, T. 2020.
\newblock MOPO: Model-based Offline Policy Optimization.
\newblock \emph{arXiv preprint arXiv:2005.13239}.

\bibitem[{Zheng, Zhang, and Grover(2022)}]{zheng2022online}
Zheng, Q.; Zhang, A.; and Grover, A. 2022.
\newblock Online decision transformer.
\newblock In \emph{International Conference on Machine Learning}. PMLR.

\end{thebibliography}

\newpage
\onecolumn
\appendix

\textbf{\LARGE Appendix}
\section*{Organization of Appendix}
The Appendix is organized as follows: \supp{sec:app_babyai} provides implementation details for the \baby environment. This is followed by additional results and discussions in \supp{sec:supp_discussion}. Ablation and comparative studies on objective and optimization methods are presented in \supp{sec:compare}. Details of the continuous tasks are outlined in \supp{sec:cont_tasks}, while \supp{sec:meta_data} includes further experiments on zero-shot adaption to different tasks in \meta. Finally, \supp{sec:rate_mat} briefly discusses sampling and the choice of the rate matrix in Algorithm~\ref{alg:sampling}.

\section{\baby Implementation Details}
\label{sec:app_babyai}
\subsection{Input and Networks.}
The various inputs and their corresponding dimensions are presented in Table~\ref{tab:input_components} for the \baby environments. We construct \model following the transformer architecture (with bi-directional masks) described in~\cite{chi_diffusion_2023}. The inputs to the model include token embeddings for observations, states, and actions, which are generated through specific mappings. Detailed illustrations of these mappings can be found in Figures~\ref{fig:method} and~\ref{fig:sample_overview}.
\begin{equation}
    \film (I_{\ctx}, s_{\ctx}, \xi, t, k) \rightarrow \tok_{\vo} \quad
    f_{\theta_s} (\vs^t_k) \rightarrow \tok_{\vs} \quad
    f_{\theta_a} (\va^t_k) \rightarrow \tok_{\va}
    \nonumber
\end{equation}
The observation token $\tok_{\vo}$ is obtained from the panoramic image $I$, state $s_{\ctx}$, instruction $\xi$, and timestep $t$ in the \gls{ctmc}. The \film (Feature-wise Linear Modulation) function~\cite{perez2018film} transforms these inputs into an observation embedding. Note that, in the case of a masked or corrupt state, these observation tokens are set to zero during training. We follow the same instruction encoder as used in the BabyAI agent model~\cite{babyai_iclr19}.
The context length, $ctx$ (see~\tbl{tbl:babyai_hyper}), acts as a memory for the agent and provides past trajectories of length $ctx$. This embedding, along with the corrupted state $\vs^t_k$ and action $\va^t_k$ tokens generated by the embedding networks $f_{\theta_s}$ and $f_{\theta_a}$, is fed into the transformer's decoder stack. The transformer's cross-attention mechanism is employed in this joint denoising process.
The decoder then predicts the true labels $\vs^1$, $\vg^1$, and $\va^1$, with the observation tokens guiding the denoising process (\fig{fig:method}). Consequently, we have $H + 1$ tokens in the cross-attention module, where the $+1$ corresponds to the \gls{ctmc} timestep-based embedding, and $2H$ tokens serve as input to the transformer's decoder block. Further details on the input dimensions and the architecture are provided in~\tbl{tbl:babyai_hyperparameters}.

\begin{table*}[ht]
\centering
\small{%
\begin{tabular}{lll}
\toprule
\textbf{Input} & \textbf{Vector Space} & \textbf{Description} \\ 
\midrule
Image $I$ & $\mathbb{R}^{S \times S \times 3}$ & Symbolic representation of env. S (size of env.)\\ 
State $s^t_k$ & $\mathbb{R}^3$ & Tuple corresponding to $<x, y, \text{dir}>$ \\ 
Goal $g^t_k$ & $\mathbb{R}^{\Pi \times 2}$ &  No. of goals ($\Pi$), each with $x, y$ positions \\ 
Instructions $\xi$ & --- & Task-specific instructions \\
Actions $a^t_k$ & $\mathbb{R}^6$ & Discrete set of actions available \\
\bottomrule
\end{tabular}%
}
\caption{\textbf{Offline data.}{ Various components of the trajectory.}}
\label{tab:input_components}
\end{table*}

\begin{table*}[ht]
\begin{center}
\begin{small}
\begin{tabular}{ll}
\toprule
\textbf{Hyperparameter} & \textbf{Value}  \\
\midrule
Number of layers & $4$  \\ 
Number of attention heads    & $4$  \\
Embedding dimension    & $128$  \\ 
Batch size   & $64$\\ 
Image Encoder & nn.Conv2d\\
Image Encoder channels & $128, 128$\\
Image Encoder filter sizes & $2 \times 2, 3 \times 3$\\
Image Encoder maxpool strides & $2, 2$ (Image Encoder may vary a little \\ & depending on the environment size)\\
Instruction Encoder & nn.GRU\\
Instruction Encoder channels & $128$\\
State Encoder & nn.Linear\\
State Encoder channels & $128, 128, 128$\\
Max epochs & $400$ \\
Dropout & $0.1$ \\
Learning rate & $8*10^{-4}$ \\
Adam betas & $(0.9, 0.95)$ \\
Grad norm clip & $1.0$ \\
Weight decay & $0.1$ \\
Learning rate decay & Linear warmup and cosine decay (see code for details) \\
\bottomrule
\end{tabular}
\caption{{Hyperparameters of \model for BabyAI experiments.}}
\label{tbl:babyai_hyperparameters}
\end{small}
\end{center}
\end{table*} 

\subsection{Environments.}

\label{sec:baby_env}

\begin{figure*}
    \centering
    \includegraphics[width=0.79\columnwidth]{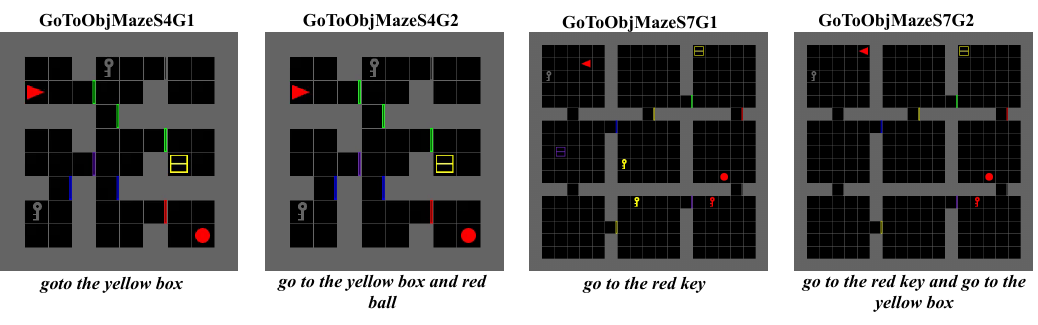}
\caption{\small{\textbf{Trajectory Planning (TP).} The agent is randomly initialized and must navigate through a maze-like environment to reach the goal(s). In each evaluation, the map layout, the agent's initial position, and goal positions are varied. Note that in the case of \model, the agent does not have access to the ground truth goal positions, whereas other baselines do.}}
    \label{fig:tp}

    \includegraphics[width=0.79\columnwidth]{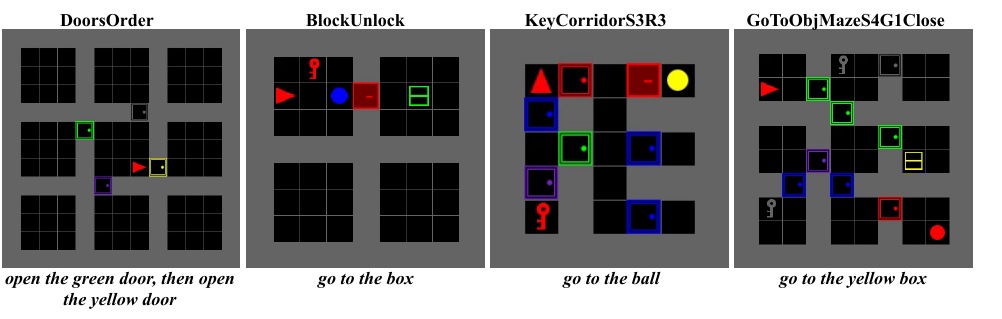}
\caption{\small{\textbf{Instruction Completion \gls{ic}.} These tasks are more complex than \gls{tp}, requiring the agent to perform additional actions such as \textit{pickup}, \textit{drop}, and \textit{open} in addition to navigation. For example, in \textbf{\textit{DoorsOpen}}, the agent must open the specified doors in the correct order without accidentally entering other rooms.}}
    \label{fig:ic}

    \includegraphics[width=\columnwidth]{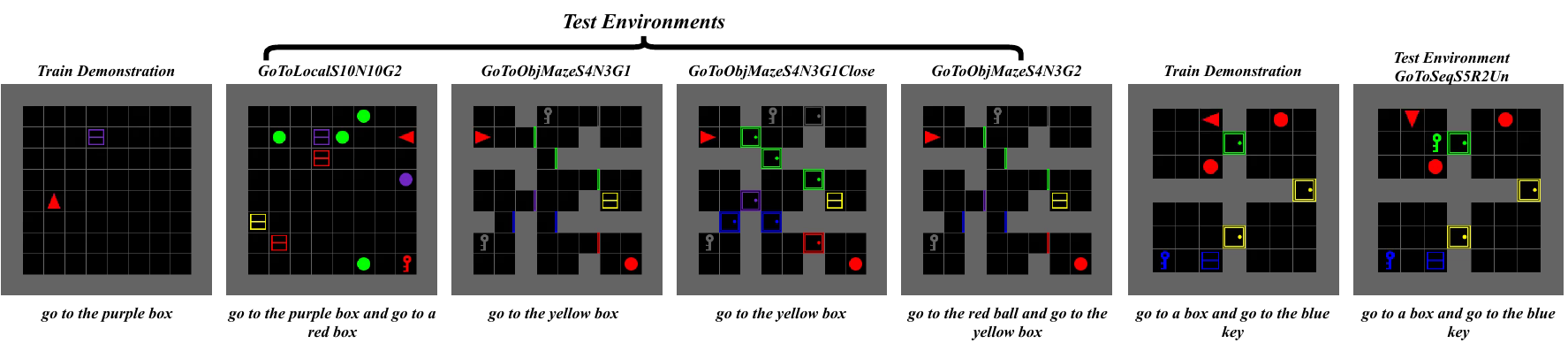}
\caption{\small{\textbf{Adaptive Planning \gls{ap}.} The model is trained on simpler tasks and tested for adaptability by progressively increasing task difficulty. In \textbf{\textit{GotoObjMazeClose}} and \textbf{\textit{GoToSeqUn}}, we evaluate whether the model can demonstrate novel skills, such as unblocking obstacles and opening doors.}}
    \label{fig:ap}

\end{figure*}

Detailed settings for the BabyAI environments are provided in Table~\ref{tbl:babyai_env_setting}. We evaluate across three task paradigms as described in~\sect{sec:babyai_exps}, with tasks in \gls{ic} and \gls{ap} being particularly challenging. For instance, in the \textbf{\textit{BlockUnlock}} task, the agent must first remove an obstacle by picking it up and repositioning it, then retrieve a key to unlock the door, and finally navigate to the goal, which is always located in the locked room. Due to the order-critical nature of these tasks, we observe that only \model succeeds, while other baselines fail. Similarly, in \gls{ap} tasks, the agent must generalize to more complex environments than those encountered during training. For further details, refer to Figures~\ref{fig:tp}, \ref{fig:ic}, and~\ref{fig:ap}.

\begin{table*}[tb]
\small
\centering
\scalebox{0.9}{
\begin{tabular}{lrrrrrrr}
\toprule
\multicolumn{1}{c}{\bf Task} & \multicolumn{1}{c}{\bf Env} & \multicolumn{1}{c}{\bf Size} & \multicolumn{1}{c}{\bf \# Room} & \multicolumn{1}{c}{\bf \# Obs} & \multicolumn{1}{c}{\bf Door} & \multicolumn{1}{c}{\bf Unblocking} & \multicolumn{1}{c}{\bf Max Steps} \\ 
\midrule

\multirow{4}{*}{\textbf{Task Planning (\gls{tp})}} 
& GoToObjMazeS4G1 & $10\times10$ & 9 & 4  & Open & No & 399 \\ 
& GoToObjMazeS4G2 & $10\times10$ & 9 & 4 & Open & No & 499 \\
& GoToObjMazeS7G1 & $19\times19$ & 9 & 7 & Open & No & 699 \\ 
& GoToObjMazeS7G2 & $19\times19$ & 9 & 7 & Open & No & 999 \\
\midrule

\multirow{4}{*}{\textbf{Instruction Completion (\gls{ic})}} 
& GoToObjMazeClose & $10\times10$ & 9 & 4 & Closed & No & 499 \\ 
& DoorsOrder & $13\times13$ & 9 & 0 & Closed & No & 144 \\ 
& BlockUnlock & $8\times8$ & 2 & 2 & Closed & Yes & 399 \\
& KeyCorridorS3R3 & $7\times7$ & 6 & 2 & Closed & No & 499 \\ 
\midrule

\multirow{4}{*}{\textbf{Adaptive Planning (\gls{ap})}} 
& GoToLocalS10N10G2 & $10\times10$ & 10 & 4 & Open & No & 299 \\ 
& GoToObjMazeS4N3G1 & $10\times10$ & 9 & 4 & Open & No & 499 \\ 
& GoToObjMazeClose & $10\times10$ & 9 & 4 & Closed & No & 499 \\ 
& GoToSeqS5R2Un & $9\times9$ & 4 & 5 &  Closed & Yes & 499 \\ 
\bottomrule
\end{tabular}
}
\caption{{BabyAI environment setting details.}}
\label{tbl:babyai_env_setting}
\end{table*}

\begin{table*}[tb]
\centering
\small
\scalebox{0.8}{
\begin{tabular}{lcc|cc|c}
\toprule
\multicolumn{3}{c}{} & \multicolumn{2}{c}{\model} & \multicolumn{1}{c}{LEAP} \\
\multicolumn{1}{c}{\bf Env} & \multicolumn{1}{c}{\bf Context} & \multicolumn{1}{c}{\bf Plan} & \multicolumn{1}{c}{\bf Iteration} & \multicolumn{1}{c}{\bf Entropy}  & \multicolumn{1}{c}{\bf Iteration} \\ 
\midrule
GoToObjMazeS4G1 & 1 & 10 & 5 & 0.3 & 10 \\ 
GoToObjMazeS4G2   &  1 & 10 & 5 & 0.3 & 10 \\ 
GoToObjMazeS7G1    & 1 & 20 & 10 & 0.3 & 50 \\ 
GoToObjMazeS7G2 & 1 & 20 & 10 & 0.3 & 50 \\ 
GoToObjMazeClose & 1 & 10 & 5 & 0.3 & 10 \\ 
DoorsOrder & 1 & 10 & 5 & 0.3 & 10 \\ 
BlockUnlock   &  10 & 30 & 15 & 0.3 & 50 \\
KeyCorridorS3R3   &  10 & 30 & 15 & 0.3 & 50 \\
GoToLocalS10N10G2   & 1 & 10 & 5 & 0.7 & 10 \\ 
GoToObjMazeS4N3G1 &   1 & 10 & 5 & 0.7 & 10 \\ 
GoToObjMazeClose  & 1 & 10 & 5 & 0.7 & 10 \\  
GoToObjMazeS4G2 &   1 & 10 & 5 & 0.7 & 10 \\ 
GoToSeqS5R2Un    & 1 & 20 & 10 & 0.7 & 40 \\
\bottomrule
\end{tabular}
}
\caption{
{\textbf{Training Configurations.} For most tasks, planning is based solely on the current state observation and aims to plan for \(H\) steps. However, for \textbf{\textit{KeyCorridor}} and \textbf{\textit{BlockUnlock}}, which involve complex sub-tasks and require memory to track completed tasks, context is used.}}
\label{tbl:babyai_hyper}
\end{table*}

\subsection{Baseline Models.}
\label{sec:baseline_supp}
\begin{figure*}[ht]
    \centering
    
    \includegraphics[width=\textwidth]{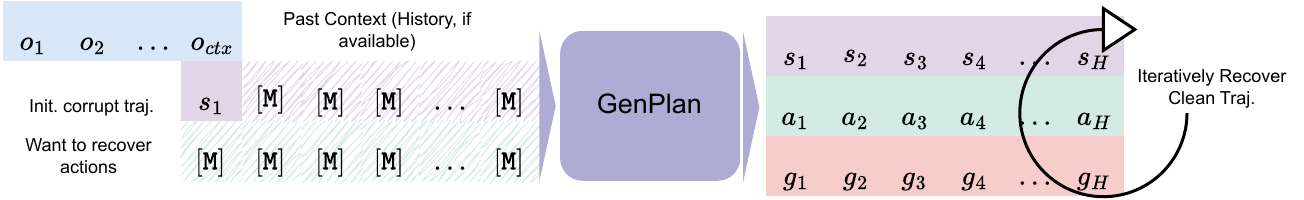}
    \caption{{\textbf{\model I/O.} During planning, the model takes in a corrupted trajectory along with the current state and past observations (if available), and iteratively recovers the clean trajectory.}}
    \label{fig:sample_overview}
\end{figure*}

\begin{figure*}
    \centering
    
    \includegraphics[width=\textwidth]{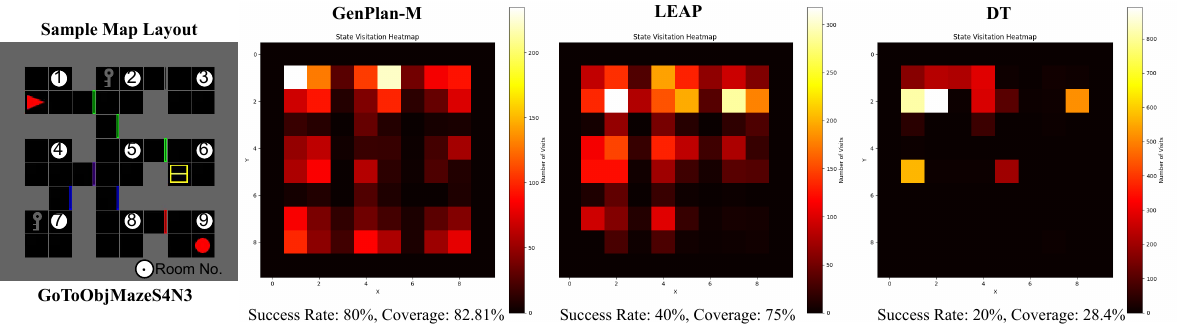}
\caption{{\textbf{State Coverage (\gls{ap}).} State visit frequency is evaluated across 10 unseen maze layouts with varying goal positions (Rooms 1-9), starting from a fixed agent position (Room 1).}}

    \label{fig:coverage_1}
    
    \includegraphics[width=\textwidth]{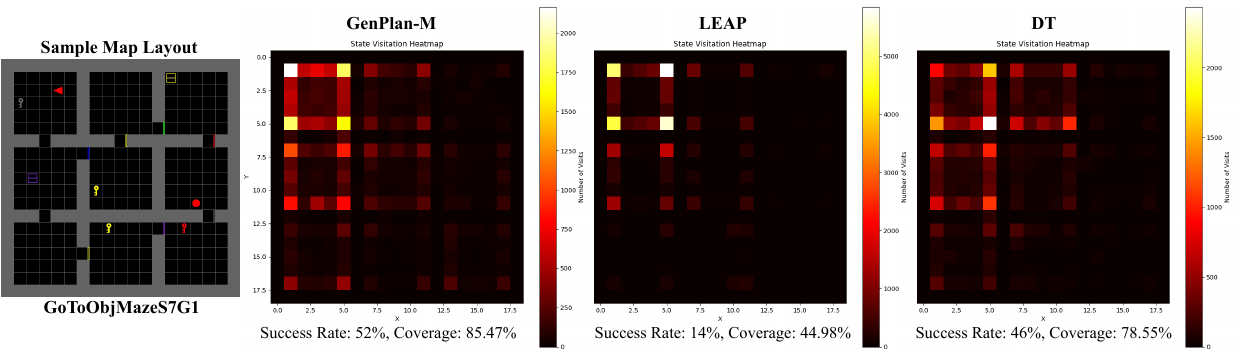}
\caption{{\textbf{State Coverage (\gls{tp}).} Coverage is evaluated in larger environments. Results are reported across 50 map variations from a fixed start position (Room 1).}}
    \label{fig:coverage_2}
    
\end{figure*}

\begin{figure*}
    \centering
        \includegraphics[width=\textwidth]{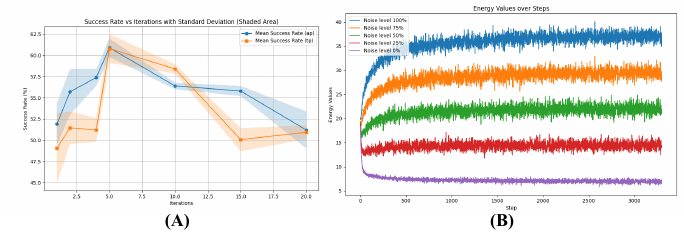}
\caption{{\textbf{\model Analysis.} \textbf{(A)} Success rate as a function of the number of denoising iterations, reported using 3 different training seeds evaluated on 250 environments, specifically for \textbf{\textit{GoToObjMazeS4N3G1 (\gls{ap})}} and \textbf{\textit{GoToObjMazeS4G1 (\gls{tp})}}. \textbf{(B)} Energy landscape learned by \model during training, illustrating that \model effectively captures the true action distribution by assigning low energy to noise-free sequences.}}
        \label{fig:add_disc_1}
\end{figure*}

\begin{figure*}
    \centering
        \includegraphics[width=\textwidth]{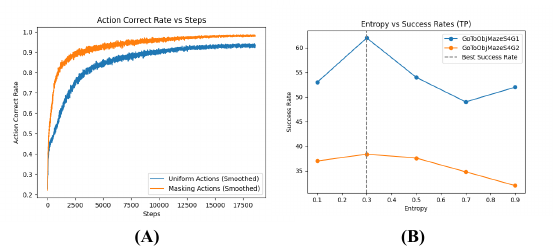}
\caption{{\textbf{\model Configurations.} \textbf{(A)} Training progression of \model, showing that \model-M converges faster than \model-U. \textbf{(B)} Success rate as a function of entropy.}}
        \label{fig:add_disc_2}
\end{figure*}

We use the official implementations for both LEAP~\cite{chen_planning_2023} and \gls{dt}~\cite{chen_decision_2021}. To address the limitations of their maximum likelihood objective, which fails to generalize in predicting novel goals (e.g., consistently predicting goals as the corner cells), we augment these baselines with simulator-based goal conditioning. The \baby simulator is used to obtain the final goal and sub-goals for goal-conditioned planning in the baseline models. All models are trained in an end-to-end fashion.

\section{Additional Results and Discussion.}
\label{sec:supp_discussion}

We present additional experiments to further evaluate \model's capabilities across various scenarios. These include analyses of coverage, iterative denoising, energy landscapes, convergence rates, and entropy, along with performance in stochastic environments and sub-optimal data. These experiments provide additional insights into \model's generalizability, robustness, and ability to adapt to diverse and complex tasks.

\subsubsection{Coverage.} 

We perform coverage experiments on both adaptive planning (\gls{ap}) and trajectory planning (\gls{tp}) tasks in novel environments. For both tasks, we evaluate models across unseen environments where the map layout (fixed) was not observed during training, and the goal position is randomized to different rooms. The results, shown in~\fig{fig:coverage_1}, report the coverage and success rates for each model.

In \gls{ap} tasks, where both the task and environment are novel, the agent is trained on a planar goal-conditioned task (without walls) and must adapt to a maze structure in a zero-shot setting. As illustrated in~\fig{fig:coverage_1}, \model significantly outperforms the baselines due to its goal generation module (\sect{sec:joint_model}), which effectively proposes sub-goals and goal regions, providing clear guidance. In contrast, baseline models struggle with ambiguous guidance (\fig{fig:goal_cond}), often leading to agent stalling, even with access to the simulator for ground truth. \model demonstrates efficient exploration and strong generalizability in these settings.

For \gls{tp} tasks, we evaluate the models' trajectory planning capabilities in a maze-goal-reaching task with unseen maps (\fig{fig:coverage_2}). This setup excludes goal modules (learned or oracle) and focuses solely on trajectory planning. Given the map and instructions, \model successfully plans trajectories, even when initializing goals in different rooms. In comparison, baseline models struggle, with \gls{dt} performing competitively in generalizing to the same task but facing significant challenges in adaptation. \model consistently outperforms baselines in both generalization and adaptation tasks.

\subsubsection{Effect of Iterative Denoising.}
At inference, the number of denoising steps for planning is determined independent of training (Algorithm~\ref{alg:sampling}), enabling post-hoc adaptations with various rate samplers. For harder problems, more iterations or better samplers can be used to obtain minimal energy trajectories. From \fig{fig:add_disc_1}A, we observe that around $H/2$ iterations provide good results, allowing approximately two dimensions to transition simultaneously in a single step.

\subsubsection{Energy Landscape.}
We investigate the energy assignment to trajectories with varying noise levels throughout training and observe that (a) \model learns to reduce the energy for clean sequences and (b) increases the energy for corrupted trajectories ({\fig{fig:add_disc_1}B}). Since this denoising is done jointly, the energy measure provides an implicit sense of the goodness of an action at a particular state (also see~\fig{fig:goal_cond}). This helps verify that the objective in \eqn{eqn:energy} indeed facilitates denoising. 

\subsubsection{Rate of Convergence \gls{dfm}.}
In the masking interpolant, the model easily identifies which indices are corrupted, simplifying the training process. This is evident through faster convergence rates. We report the correct action rate (i.e., the portion of the corrupted trajectory that the model can recover) and observe that masked denoising converges much faster (\fig{fig:add_disc_2}A). Consequently, the masked tokens receive the lowest attention scores. In contrast, it is relatively harder for the model to determine which tokens are corrupted in the uniform interpolant, leading to slower convergence.

\subsubsection{Entropy.}
Continuing our discussion from \sect{sec:babyai_exps}, we report success rates by varying the entropy lower bound. For \gls{tp} and \gls{ic} tasks, we observe that entropy levels above 0.3 offer little benefit. However, for \gls{ap} tasks, we set the entropy to 0.7, as this facilitates task discovery (i.e., performing novel actions that are absent from the dataset) (see \fig{fig:add_disc_2}B).

We extend the results presented in~\tbl{tbl:adaptive_comparison} by introducing an additional variant, \textbf{LEAP$\oplus\mathcal{H}$}. Similar to \model, this variant of LEAP incorporates a lower bound on entropy, facilitating novel task adaptation.

\begin{table}[H]
\centering
\small
\begin{tabular}{|l|c|c|c|c|c|c|}
    \hline
    & \multicolumn{4}{c|}{\textbf{Unconditional Rollouts}} & \multicolumn{2}{c|}{\textbf{Conditional Rollouts}} \\
    \hline
    \textbf{Environment} & \textbf{\model-U} & \textbf{\model-M} & \textbf{LEAP$\ominus$GC} & \textbf{LEAP$\oplus\mathcal{H}$} & \textbf{LEAP} & \textbf{DT} \\
    \hline
    \multicolumn{7}{|c|}{\textbf{Adaptive Planning (AP)}} \\
    \hline
    GoToLocalS10N10G2 & 82.4\% & \textbf{88\%} & 76\% & 69.2\% & 78\% & 25.6\% \\
    GoToObjMazeS4N3G1 & 56\% & \textbf{62\%} & 44.8\% & 52\% & 48\% & 24\% \\ 
    GoToObjMazeClose & 31.2\% & \textbf{34.8\%} & 10\% & 16.4\% & 10\% & 8.8\% \\
    GoToObjMazeS4G2 & 28.8\% & \textbf{34.8\%} & 14\% & 21.2\% & 18.4\% & 3.6\% \\ 
    GoToSeqS5R2Un & 35.6\% & \textbf{42\%} & 29.2\% & 30.8\% & 38\% & 29.2\% \\
    \hline
    \textbf{AP Mean (13.84 $\uparrow$)} & 46.8\% & \textbf{52.32\%} & 34.8\% & 37.92\% & 38.48\% & 18.24\% \\
    \hline
\end{tabular}
\caption{{\textbf{\baby quantitative performance.} Success rates of the models across different environments are presented. The abbreviations \textbf{SW}, \textbf{NX}, \textbf{RY}, and \textbf{GZ} in the environment names represent the size (\textbf{W}) of a room in the map, the number of obstacles (\textbf{X}), the number of rows (\textbf{Y}), and the number of goals (\textbf{Z}) during testing, respectively. See \supp{sec:baby_env} for more details.}}
\label{tab:final}
\end{table}

\subsubsection{Stochasticity in Environments.} 
In this setup, the agent operates in an environment where it has a 20\% chance that its chosen action (e.g., left or right) will be randomly mapped to another action with uniform probability. To address these stochastic settings, we replan after each step using single-step rollouts, unlike the multi-step rollouts used in the settings presented in Tables~\ref{tbl:performance_comparison} and~\ref{tbl:adaptive_comparison}. We apply the same replanning strategy to the baseline models for a fair comparison.

The results in~\tbl{tbl:stoch_exps} show that \model performs on par with baseline methods in stochastic environments, where both use feedback at each step. However, in adaptive environments, even with feedback, the baseline model (\gls{dt}) fails, while \model and LEAP succeed.

\begin{table}[H]
\small
\centering
\begin{tabular}{|c|c|c|c|}
\hline
\textbf{Env} & \textbf{GenPlan-M} & \textbf{LEAP} & \textbf{DT} \\ \hline
GoToObjMazeS7G1 (\gls{tp}) & 48\% & 18\% & \textbf{52\%} \\ \hline
GoToObjMazeS4N3G1 (\gls{ap}) & \textbf{54.8\%} & 52\% & 23.2\% \\ \hline
\end{tabular}
\caption{{\textbf{Stochastic Environments.} Success rates of models in task planning and adaptive environments using single-step rollouts to account for stochasticity.}}
\label{tbl:stoch_exps}
\end{table}

\subsubsection{Context Length.}
We use context length as a memory mechanism for tracking completed sub-goals, which is primarily beneficial in tasks like \textbf{KeyCorridorS3R3} and \textbf{BlockUnlock} (\tbl{tbl:performance_comparison}). While most \gls{tp} tasks and multi-objective \gls{ic} tasks operate under a Markovian assumption, these tasks involve complex sub-tasks requiring memory to track progress. Context-based planning is also enabled for baselines (see~\fig{fig:sample_overview})

\subsubsection{Time considerations.}
We observe that due to the \gls{mlm} objective in LEAP, they must sequentially mask each action, which is time-consuming. Additionally, gradients are updated only after accumulating losses from all action tokens in the horizon, making the process memory-intensive. In contrast, \model controls the number of masked tokens using the \gls{ctmc} timestep $t$, eliminating the need for sequential masking and saving significant time. Although \model is slightly slower than \gls{dt} due to its iterative inference process and the bi-directional mask loss taking longer than the causal masks in \gls{dt}, it still performs within a reasonable time frame. The total training and inference times are reported in \tbl{tbl:times}.
\begin{table}[H]
\small
\centering
\begin{tabular}{|c|c|c|c|c|}
\hline
\textbf{Mode} & \textbf{Env} & \textbf{GenPlan-M} & \textbf{LEAP} & \textbf{DT} \\ \hline
\multirow{2}{*}{\textbf{Train}} & GoToObjMazeS4G2 & 34m & 5hrs & 30min \\ \cline{2-5}
                                & GoToObjMazeS7G2 & 1hr 3min & 6hrs & 57min \\ \hline
\multirow{2}{*}{\textbf{Eval}} & GoToObjMazeS4  & 3mins & 10min & 2min \\ \cline{2-5}
                                & GoToObjMazeS7 & 18min & 36min & 10min \\ \hline
\end{tabular}
\caption{{Comparison of training and evaluation times across different environments and methods. Training is performed with a batch size of 64 over 500 epochs, while evaluation is conducted across 250 environments.}}
\label{tbl:times}
\end{table}

\subsubsection{Instructions Ablation.}  
We observe that image inputs play a significant role in learning goal distributions. The training data comprises standalone examples of single goals or skills, while testing involves skill chaining based on prompts in a fully observable environment. GenPlan addresses these challenging tasks by learning energy landscapes and employing DFM sampling.

Prompts are essential in instruction-completion tasks. Procedurally generated by MiniGrid environments, these prompts are available during both training and testing phases. To evaluate their significance, we ablate the instruction embeddings, which are key to instruction completion as seen in~\tbl{table:success_rates_reduction}. While the primary focus of this work is to adopt discrete representations for planning with sub-tasks, language grounding is left for future exploration.

\begin{table}[h!]
\centering
\begin{tabular}{|l|c|c|}
\hline
\textbf{Environment} & \textbf{Success Rate} & \textbf{Reduction} \\ \hline
Doors Order (IC) & 22\% & 13.2\% $\downarrow$ \\ \hline
GoToLocal10G2 (AP) & 73\% & 15\% $\downarrow$ \\ \hline
\end{tabular}
\caption{{\textbf{Instruction Prompt Ablation.} 
We ablate the instruction embeddings from the input to assess their significance, as evidenced by the reduction in success rates across environments.}}
\label{table:success_rates_reduction}
\end{table}

\subsubsection{Suboptimal Data.}  
We evaluate the performance of our model when trained on suboptimal data and observe that \model-M retains its generalization capability, while DT fails to adapt effectively. We hypothesize that this is due to the MLE objective in DT being unable to stitch trajectories.

\begin{table}[h!]
\centering
\begin{tabular}{|c|c|c|c|c|}
\hline
\multirow{2}{*}{\textbf{Environment}} & \multicolumn{2}{c|}{\textbf{Optimal Data}} & \multicolumn{2}{c|}{\textbf{Suboptimal Data (25\%)}} \\ \cline{2-5} 
 & \textbf{\model-M} & \textbf{DT} & \textbf{\model-M} & \textbf{DT} \\ \hline
GoToObjMazeS4G1 & \textbf{62\%} & 44\% & \textbf{55.2\%} & 44\% \\ \hline
GoToObjMazeS7G1 & \textbf{44.8\%} & 40\% & \textbf{38.8\%} & 15.2\% \\ \hline
BlockUnlock & \textbf{16\%} & 0\% & \textbf{8\%} & 0\% \\ \hline
\end{tabular}
\caption{{\textbf{Performance with Suboptimal Trajectories.}  
We introduce 25\% random actions into the training data to simulate suboptimal trajectories. Despite this, \model-M demonstrates robust generalization, outperforming DT across all environments.}}
\label{table:genplan_performance}
\end{table}

\section{Comparisons with Generative Approaches}
\label{sec:compare}

We evaluate different generative objectives (\gls{bc}, Noise Contrastive Estimation) and sampling techniques (Random Shooting (RS), Cross-Entropy Method (CEM), MCMC sampling, and energy gradient-guided sampling). The results for the \gls{ap} and \gls{ic} tasks are reported in~\tbl{table:performance_comparison}.

(i) \textbf{Denoising Diffusion Probabilistic Models (DDPM)}~\cite{ho2020denoising, chi_diffusion_2023} rely on a \gls{bc} objective thus struggle to generalize to harder tasks, as shown in Example~\ref{ex:adapt}. Models trained solely with \gls{bc} objectives lack distributional fidelity, limiting their ability to adapt to novel scenarios.

(ii) \textbf{Energy models} are challenging to sample from due to their reliance on extensive CEM or MCMC cycles and their instability during training~\cite{chi_diffusion_2023}. Additionally, guided sampling in DDPM (using energy gradients) often results in local minima, where low-energy trajectories are gamed through repeating actions frequently observed in the dataset (e.g., repeated forward movements). 

We observe that \model significantly outperforms both approaches, as demonstrated in~\tbl{table:performance_comparison}.

\begin{table}[h!]
\centering
\begin{tabular}{|l|l|c|c|c|}
\hline
\multicolumn{2}{|c|}{\textbf{Sampling / Objective}} & \textbf{GoToObjMazeS4G1} & \textbf{KeyCorridorS3R3} & \textbf{DoorsOrder} \\ \hline
\multirow{2}{*}{\textbf{Energy Models}} 
    & Random & 29.2\% & 2.8\% & 26\% \\ 
    & CEM & {38.9\%} & 6.4\% & 28.4\% \\ \hline
\multirow{2}{*}{\textbf{DDPM}} 
    & Diffusion BC & 25.2\% & 0.8\% & 29.2\% \\ 
    & Energy (Gradient) & 2\% & 0\% & 0\% \\ \hline
\textbf{GenPlan} & Energy+DFM & \textbf{62\%} & \textbf{17.6\%} & \textbf{35.2\%} \\ \hline
\end{tabular}
\caption{\textbf{Comparison of Energy-Based and DDPM Baselines on MiniGrid Tasks.}  
We compare various sampling techniques and generative objectives on discrete planning tasks. GenPlan demonstrates significant improvements in success rates, particularly on more complex tasks, by combining energy objectives with DFM sampling.}
\label{table:performance_comparison}
\end{table}

\section{Continuous Tasks Implementation Details.}
\label{sec:cont_tasks}

\subsubsection{Simulation Setup.}
(i) \textbf{PushT}~\cite{florence2021implicit}: The goal is to push a T-block to a target position and orientation using an end-effector controlled via 2D velocity. The dataset includes 206 human demonstrations.
(ii) \textbf{Kitchen}~\cite{gupta2019relay}: A robotic manipulation task with a Franka Panda arm (7D action space) and 566 demonstrations. It comprises 7 sub-tasks, with each trajectory completing 4 tasks in some order. At test time, the metric is the number of tasks completed in an unconditional rollout. Both tasks are evaluated in state space variants.

\begin{table}[h!]
\small
\centering
\begin{minipage}{0.25\textwidth}
\centering
\includegraphics[width=0.6\textwidth]{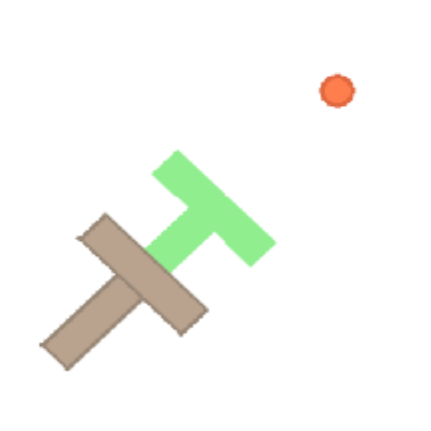} %
\caption{{\textbf{PushT env.}}}
\label{fig:image}
\end{minipage}%
\hfill
\begin{minipage}{0.75\textwidth}
\centering
\begin{tabular}{|c|c|c|c|c|c|}
\hline
\textbf{Env} & \textbf{Metric} & \textbf{\model-M} & \textbf{VQ-BeT} & \textbf{DP-C} & \textbf{DP-T}  \\ \hline
\multirow{2}{*}{PushT} & Final Coverage & {0.73} & 0.7 & 0.73 & \textbf{0.74} \\ 
 & Max Coverage & {0.77} & 0.73 & \textbf{0.86} & 0.83 \\ \hline
 Kitchen & \# Tasks &  3.40 & \textbf{3.66} & 2.62 & 3.44 \\
 \hline

\end{tabular}
\caption{{\textbf{PushT Task Evaluation.} Comparison of state-based and image-based policies.}}
\label{tbl:pusht}
\end{minipage}
\end{table}

Evaluation metrics include final and maximum coverage, measured by the Intersection over Union (IoU) between the T-block and the target T position.

We compare our approach with VQ-BeT~\cite{lee2024behavior}, which employs a categorical transformer-based model to infer action codes, subsequently decoded using a pre-trained motion primitive model (VQ-VAE). In our experiments, we use the same pre-trained motion primitive but replace the auto-regressive transformer with \model's objective and sampling process.

\section{\meta Implementation Details.}
\label{sec:meta_data}

We assess the adaptive capabilities of the agents in \meta's unseen tasks.

\subsubsection{Simulation Setup.}
Our simulation setup closely follows that of~\cite{schmied2023learning}, benchmarking \model on \meta tasks that encompass various robotic manipulation challenges. Both MDDT~\cite{schmied2023learning} and \model are trained on five tasks (In-Dist.) and evaluated on nine tasks, including four new tasks (Out-Dist.) designed to test adaptation. Consistent with the \baby experiments, we directly evaluate zero-shot performance. Each episode consists of 200 timesteps per task, and we report success rates and mean rewards across Meta-World experiments.

To enable multi-domain learning and task adaptation, the action space is modeled as a categorical distribution with actions discretized into 64 bins using min-max tokenization, as in~\cite{schmied2023learning}. In \meta, the state and action spaces are consistent across environments ($|S| = 39, |A| = 4$). The dataset comprises 10K trajectories per task (length 200) collected by task-specific \gls{sac} agents, featuring behavior ranging from random to expert-level. Unlike the near-optimal demonstrations in \baby, this dataset offers broader variability. Sequential action prediction with a planning horizon of 1 is employed, which provides feedback at each step and improves adaptive performance. The environments are detailed in~\tbl{tbl:mt40}.

\begin{table}[h!]
\centering
\begin{small}
\begin{tabular}{l l r r}
    \toprule
    \textbf{Category} & \textbf{Metric} & \textbf{MDDT} & \textbf{\model-M} \\
    \midrule
    \multirow{2}{*}{\textbf{In-Dist.}} 
        & \textbf{SR} & \textbf{0.88 $\pm$ 0.032} & 0.786 $\pm$ 0.104 \\
        & \textbf{MR} & \textbf{1566 $\pm$ 38.0} & 1465.66 $\pm$ 34.79 \\
    \midrule
    \multirow{2}{*}{\textbf{Out-Dist.}} 
        & \textbf{SR} & 0.3 $\pm$ 0.081 & \textbf{0.333 $\pm$ 0.047} \\
        & \textbf{MR} & 617 $\pm$ 52.46 & \textbf{620.33 $\pm$ 48.58} \\
    \bottomrule
\end{tabular}
\end{small}
\caption{{\textbf{Quantitative Performance.} Comparison between MDDT and \model. Success rates (SR) and mean rewards (MR) are reported across three unique seeds.}}
\label{tab:quant_performance}
\end{table}

\paragraph{Rollout.}
During evaluation, actions are sequentially predicted token by token to fill the 4-dimensional action space ($|A|=4$). After completing the action, the corresponding observation is obtained (plan horizon of 1).

\subsubsection{Baselines and Results.}
\begin{table}[h!]
    \small
    \centering
    \begin{tabular}{l c c c c r r}
    \toprule
    & \multicolumn{4}{c}{Dataset} & \multicolumn{2}{c}{Evaluation - SR} \\
    \textbf{Task} & $|\mathcal{S}|$ & $|\mathcal{A}|$ & \textbf{SR} & \textbf{Reward} & \textbf{GenPlan} & \textbf{MDDT} \\
    \midrule
    \multicolumn{7}{l}{\textbf{In Dist. Environments}} \\
    button-press-v2 & 39 & 4 & 1.0 & 1430.44 & 0.73$\pm$0.11 & \textbf{0.8$\pm$0.2} \\
    dial-turn-v2 & 39 & 4 & 0.8 & 1674.29 & 0.86$\pm$0.23 & \textbf{0.93$\pm$0.11} \\
    disassemble-v2 & 39 & 4 & 1.0 & 1396.55 & 0.66$\pm$0 & \textbf{0.8$\pm$0.2} \\
    plate-slide-v2 & 39 & 4 & 1.0 & 1667.35 & \textbf{1$\pm$0} & \textbf{1$\pm$0} \\
    reach-v2 & 39 & 4 & 1.0 & 1858.99 & 0.73$\pm$0.11 & \textbf{0.933$\pm$0.11} \\
    \midrule
    \multicolumn{7}{l}{\textbf{Adaptive Environments (Out Dist.)}} \\
    reach-wall-v2 & 39 & 4 & 1.0 & 1831.14 & \textbf{0.86$\pm$0.11} & 0.73 $\pm$0.11 \\
    plate-slide-side-v2 & 39 & 4 & 1.0 & 1663.35 &  0$\pm$0 & 0$\pm$0 \\
    button-press-wall-v2 & 39 & 4 & 1.0 & 1508.16 & \textbf{0.46$\pm$0.11} & 0.4$\pm$0.2 \\
    coffee-button-v2 & 39 & 4 & 1.0 & 1499.17 & \textbf{0.13$\pm$0.067} & 0.2$\pm$0.2 \\
    \bottomrule
    \end{tabular}
    \caption{{\textbf{Dataset and Performance Scores.} The agent is trained only on datasets from the 5 in-distribution environments and evaluated on all nine. Datasets from adaptive environments are not used.}}
    \label{tbl:mt40}
\end{table}

We follow the pre-processing steps from MDDT~\cite{schmied2023learning}, with \model utilizing a flow-based generative loss instead of the auto-regressive loss, as described in \sect{sec:method}. During sampling, return-conditioned sampling~\cite{chen_decision_2021} is used. While MDDT performs well on in-distribution tasks, \model demonstrates superior generalization to closely related tasks (e.g., \textit{reach} and \textit{button} tasks) and adapts effectively to variations in orientation or axis without fine-tuning. 

The masking variant of \model benefits from a robust training process and multi-modality by predicting alternate actions at corrupt timesteps in the trajectory. As a result, \model achieves slightly better scores in adaptive tasks.

\section{Rate Matrices and Sampling}
\label{sec:rate_mat}
\begin{table}[h!]
\centering
\begin{minipage}{0.48\textwidth}
\centering
\small
\begin{tabular}{|c|c|c|c|c|c|c|c|c|c|c|}
\hline
\rowcolor{gray!20} \textbf{t} & \textbf{$k_1$} & \textbf{$k_2$} & \textbf{$k_3$} & \textbf{$k_4$} & \textbf{$k_5$} & \textbf{$k_6$} & \textbf{$k_7$} & \textbf{$k_8$} & \textbf{$k_9$} & \textbf{$k_{10}$} \\ 
\hline
\textcolor{blue}{0.0} & \cellcolor{red!30}6 & \cellcolor{red!30}6 & \cellcolor{red!30}6 & \cellcolor{red!30}6 & \cellcolor{red!30}6 & \cellcolor{red!30}6 & \cellcolor{red!30}6 & \cellcolor{red!30}6 & \cellcolor{red!30}6 & \cellcolor{red!30}6 \\ 
\hline
\textcolor{blue}{0.2} & \cellcolor{green!30}1 & \cellcolor{red!30}6 & \cellcolor{red!30}6 & \cellcolor{green!30}0 & \cellcolor{red!30}6 & \cellcolor{green!30}5 & \cellcolor{red!30}6 & \cellcolor{red!30}6 & \cellcolor{red!30}6 & \cellcolor{red!30}6 \\ 
\hline
\textcolor{blue}{0.4} & \cellcolor{green!30}1 & \cellcolor{red!30}6 & \cellcolor{red!30}6 & \cellcolor{green!30}0 & \cellcolor{red!30}6 & \cellcolor{green!30}5 & \cellcolor{red!30}6 & \cellcolor{red!30}6 & \cellcolor{green!30}2 & \cellcolor{red!30}6 \\ 
\hline
\textcolor{blue}{0.6} & \cellcolor{green!30}1 & \cellcolor{green!30}5 & \cellcolor{red!30}6 & \cellcolor{green!30}0 & \cellcolor{red!30}6 & \cellcolor{green!30}5 & \cellcolor{red!30}6 & \cellcolor{red!30}6 & \cellcolor{green!30}2 & \cellcolor{red!30}6 \\ 
\hline
\textcolor{blue}{0.8} & \cellcolor{green!30}1 & \cellcolor{green!30}5 & \cellcolor{green!30}0 & \cellcolor{green!30}0 & \cellcolor{green!30}2 & \cellcolor{green!30}5 & \cellcolor{green!30}1 & \cellcolor{green!30}1 & \cellcolor{green!30}2 & \cellcolor{green!30}2 \\ 
\hline
\textcolor{blue}{1.0} & \cellcolor{green!30}1 & \cellcolor{green!30}5 & \cellcolor{green!30}0 & \cellcolor{green!30}0 & \cellcolor{green!30}2 & \cellcolor{green!30}5 & \cellcolor{green!30}1 & \cellcolor{green!30}1 & \cellcolor{green!30}2 & \cellcolor{green!30}2 \\ 
\hline
\end{tabular}
\caption{{\textbf{Masking Interpolant.} The sample is initialized with the mask state (represented by 6) and iteratively denoised to obtain the clean sequence (in green).}}
\label{table:1}
\end{minipage}
\hfill
\begin{minipage}{0.48\textwidth}
\centering
\small
\begin{tabular}{|c|c|c|c|c|c|c|c|c|c|c|}
\hline
\rowcolor{gray!20} \textbf{t} & \textbf{$k_1$} & \textbf{$k_2$} & \textbf{$k_3$} & \textbf{$k_4$} & \textbf{$k_5$} & \textbf{$k_6$} & \textbf{$k_7$} & \textbf{$k_8$} & \textbf{$k_9$} & \textbf{$k_{10}$} \\ 
\hline
\textcolor{blue}{0.0} & \cellcolor{red!30}3 & \cellcolor{green!30}5 & \cellcolor{red!30}4 & \cellcolor{red!30}3 & \cellcolor{green!30}2 & \cellcolor{red!30}4 & \cellcolor{green!30}1 & \cellcolor{green!30}2 & \cellcolor{red!30}0 & \cellcolor{green!30}2 \\ 
\hline
\textcolor{blue}{0.2} & \cellcolor{red!30}3 & \cellcolor{green!30}5 & \cellcolor{red!30}4 & \cellcolor{green!30}0 & \cellcolor{green!30}2 & \cellcolor{red!30}4 & \cellcolor{green!30}1 & \cellcolor{green!30}2 & \cellcolor{red!30}0 & \cellcolor{green!30}2 \\ 
\hline
\textcolor{blue}{0.4} & \cellcolor{red!30}3 & \cellcolor{green!30}5 & \cellcolor{red!30}4 & \cellcolor{green!30}0 & \cellcolor{green!30}2 & \cellcolor{red!30}4 & \cellcolor{green!30}1 & \cellcolor{green!30}2 & \cellcolor{green!30}2 & \cellcolor{green!30}2 \\ 
\hline
\textcolor{blue}{0.6} & \cellcolor{green!30}1 & \cellcolor{green!30}5 & \cellcolor{green!30}0 & \cellcolor{green!30}0 & \cellcolor{green!30}2 & \cellcolor{red!30}4 & \cellcolor{green!30}1 & \cellcolor{green!30}1 & \cellcolor{green!30}2 & \cellcolor{green!30}2 \\ 
\hline
\textcolor{blue}{0.8} & \cellcolor{green!30}1 & \cellcolor{green!30}5 & \cellcolor{green!30}0 & \cellcolor{green!30}0 & \cellcolor{green!30}2 & \cellcolor{green!30}5 & \cellcolor{green!30}1 & \cellcolor{green!30}1 & \cellcolor{green!30}2 & \cellcolor{green!30}2 \\ 
\hline
\textcolor{blue}{1.0} & \cellcolor{green!30}1 & \cellcolor{green!30}5 & \cellcolor{green!30}0 & \cellcolor{green!30}0 & \cellcolor{green!30}2 & \cellcolor{green!30}5 & \cellcolor{green!30}1 & \cellcolor{green!30}1 & \cellcolor{green!30}2 & \cellcolor{green!30}2 \\ 
\hline
\end{tabular}
\caption{{\textbf{Uniform Interpolant.} The sample is initialized with uniform noise (in red) and iteratively denoised to obtain the clean trajectory (in green).}}
\label{table:2}
\end{minipage}
\end{table}

Once the denoising model $\denoise^\theta$ is trained, we can select a rate matrix to simulate the \gls{ctmc} for generating samples (see Algorithm~\ref{alg:sampling}). 
\citeauthor{campbell2024generative} proposes the following rate matrix definition as an initial choice (for simplicity, consider the case where $H=1$):
\begin{align*}
    \relurate_t(\x^t_k, j_k | \x^1_k) \vcentcolon= \frac{\relu \left( \partial_t \noisemarg(j_k | \x^1_k) - \partial_t \noisemarg(\x^t_k | \x^1_k) \right)}{\mathcal{Z}_t \noisemarg(\x^t_k | \x^1_k)},
    \label{eq:relu_rate}
\end{align*}
where $\relu(a) = \max(a, 0)$ and $\mathcal{Z}_t$ is the number of states with non-zero mass at $t$, defined as $\mathcal{Z}_t = | \{ \x^t_k : \noisemarg(\x^t_k | \x^1_k) > 0 \} |$. Note that $\relurate_t(\x^t_k, j_k | \x^1_k) = 0$ when $\noisemarg(\x^t_k | \x^1_k) = 0$ or $\noisemarg(j_k | \x^1_k) = 0$. Additionally, when $\x^t_k = j_k$, we have $\relurate_t(\x^t_k, \x^t_k | \x^1_k) = - \sum_{j_k \neq \x^t_k} \relurate_t(\x^t_k, j_k | \x^1_k)$.

To ensure that the rate matrix $\relurate_t$ is well-defined and maintains the flow of probability mass between states, we assume that $\noisemarg(\x^t_k | \x^1_k) > 0$. This assumption is crucial because it ensures that the state $\x^t_k$ is ``alive" and carries some probability mass at time $t$. If $\noisemarg(\x^t_k | \x^1_k) = 0$, then $\relurate_t(\x^t_k, j_k | \x^1_k)$ would either be undefined or forced to be zero, as the state would be considered ``dead," unable to either receive or provide any mass. Without this assumption, a state could demand incoming mass without being able to reciprocate, violating the consistency of the conditional flow process and the Kolmogorov equation.

\end{document}